\documentclass[preprint,11pt,review]{elsarticle}

\usepackage{graphicx}
\usepackage[dvipsnames]{xcolor}
\usepackage{caption}
\usepackage{subcaption}
\usepackage[makeroom]{cancel}
\usepackage{booktabs}
\usepackage{graphics}
\usepackage{adjustbox}
\usepackage{float}
\usepackage{multirow}
\usepackage{url}
\usepackage{color}

\usepackage[title]{appendix}

\usepackage[linesnumbered,ruled,vlined]{algorithm2e}
\SetKwInput{KwInput}{Input}
\SetKwInput{KwOutput}{Output}

\graphicspath{{./img/}}

\usepackage{hyperref}
\usepackage[utf8]{inputenc}

\usepackage{bm}
\usepackage{amssymb}

\usepackage{amsmath}

\def\R{{\mathbb R}}

\def\p{{\mathrm{p}}}
\def\q{{\mathrm{q}}}
\def\dd{{\mathrm{d}}}

\def\Q{{\mathrm{Q}}}

\def\const{{\mathrm{const}}}

\def\Ebb{{\mathbb{E}}}

\def\diag{\operatorname{Diag}}

\def\PG{\operatorname{PG}}
\def\logit{\operatorname{logit}}

\def\bzero{{\mathbf 0}}
\def\bone{{\mathbf 1}}

\def\blambda{{\boldsymbol{\lambda}}}
\def\bmu{{\boldsymbol{\mu}}}

\def\bpi{{\boldsymbol{\pi}}}

\def\bomega{{\boldsymbol{\omega}}}

\def\bTheta{{\boldsymbol{\Theta}}}

\def\bLambda{{\boldsymbol{\Lambda}}}

\def\bOmega{{\boldsymbol{\Omega}}}

\def\ba{{\mathbf a}}

\def\bff{{\mathbf f}}

\def\bm{{\mathbf m}}

\def\bu{{\mathbf u}}

\def\bx{{\mathbf x}}
\def\by{{\mathbf y}}
\def\bz{{\mathbf z}}

\def\bI{{\mathbf I}}

\def\bK{{\mathbf K}}

\def\bS{{\mathbf S}}
\def\bT{{\mathbf T}}

\def\bX{{\mathbf X}}

\def\bZ{{\mathbf Z}}

\usepackage{mathtools}

\usepackage{amsthm}
\usepackage[createShortEnv]{proof-at-the-end}

\usepackage{xcolor}
\colorlet{review}{black}

\journal{Artificial Intelligence Journal}

\begin{document}

\begin{frontmatter}



\title{Hyperbolic Secant Representation of the Logistic Function: Application to Probabilistic Multiple Instance Learning for CT Intracranial Hemorrhage Detection}


\author[CCIA,CITIC]{Francisco M. Castro-Macías\corref{cor1}}
\ead{fcastro@ugr.es}
\author[STAT,CITIC]{Pablo Morales-Álvarez}
\author[NWU]{Yunan Wu}
\author[CCIA]{Rafael Molina}
\author[NWU,MED]{Aggelos K. Katsaggelos}
\address[CCIA]{Department of Computer
	Science and Artificial Intelligence, University of Granada, Spain.}
\address[CITIC]{Research Centre for Information and Communication Technologies, University of Granada, Spain.}
\address[STAT]{Department of Statistics and Operations Research, University of Granada, Spain.}
\address[NWU]{Department of Electrical and Computer
Engineering, Northwestern University, USA.}
\address[MED]{Center for Computational Imaging and Signal Analytics in Medicine, Northwestern University, USA.}
\cortext[cor1]{Corresponding author.}
\date{}

\begin{abstract}
Multiple Instance Learning (MIL) is a weakly supervised paradigm that has been successfully applied to many different scientific areas and is particularly well suited to medical imaging. 
Probabilistic MIL methods, and more specifically Gaussian Processes (GPs), have achieved excellent results due to their high expressiveness and uncertainty quantification capabilities. 
One of the most successful GP-based MIL methods, VGPMIL, resorts to a variational bound to handle the intractability of the logistic function. 
Here, we formulate VGPMIL using Pólya-Gamma random variables. 
\textcolor{review}{This approach yields the same variational posterior approximations as the original VGPMIL, which is a consequence of the two representations that the Hyperbolic Secant distribution admits.}
This leads us to propose a general GP-based MIL method that takes different forms by simply leveraging distributions other than the Hyperbolic Secant one. 
Using the Gamma distribution we arrive at a new approach that obtains competitive or superior predictive performance and efficiency. 
This is validated in a comprehensive experimental study including one synthetic MIL dataset, two well-known MIL benchmarks, and a real-world medical problem. 
We expect that this work provides useful ideas beyond MIL that can foster further research in the field.
\end{abstract}



\end{frontmatter}



\section{Introduction}

Multiple Instance Learning (MIL) \cite{CARBONNEAU2018329} is a type of weakly supervised learning that has become very popular due to the reduced annotation effort it requires. 
In MIL binary classification \cite{dietterich1997solving} the training set consists of instances grouped into bags. 
Both bags and instances have labels, but we only observe them at the bag level while instance labels remain unknown. 
It is assumed that a bag label is positive if and only if it  contains at least one positive instance. 
The goal is to achieve a method capable of accurately predicting both bag and instance labels using only bag labels.

The MIL approach has been successfully applied to many different scientific domains \cite{CARBONNEAU2018329}, being particularly well suited to medical imaging \cite{quellec2017multiple}.
In this work, we are particularly interested in the problem of IntraCranial Hemorrhage (ICH) detection. 
ICH is a severe life-threatening emergency with high mortality and morbidity rates caused by blood leakage inside the brain \cite{otite2017ten}. 
Computed Tomography (CT) scans are widely used to diagnose ICH because it is an inexpensive and non-invasive technique for patients. 
Each scan is made up of a significant number of slices, each representing a section of the head at a given height. 
A CT scan is labeled as ICH if at least one of its slices shows evidence of the injury, otherwise, it is normal. 
To apply a supervised learning approach, radiologists have to manually label every single slice in the dataset \cite{phong2017brain, chilamkurthy2018deep, cho2019improving}, which is a costly and tedious process. 
In contrast, the MIL approach significantly reduces the radiologists' workload as only one label for each scan is needed. 

To learn in the MIL scenario, Deep Learning (DL) methods have become popular in practice due to their ability to deal with highly structured data \cite{quellec2017multiple, saab2019doubly, remedios2020extracting, qi2021dr}
The most successful models combine DL architectures with attention mechanisms to weigh the relevance of each instance, see \cite{ilse2018attention}. 
However, these methods do not model the instance label \emph{explicitly} (they just have an attention weight for each instance), which hampers the quantification of uncertainty at the instance level. 
Note that this is essential in MIL since instance labels are unknown.
As a consequence, plenty of attention has been paid to probabilistic MIL methods in recent years. 
Among them, Gaussian Processes (GPs) have achieved very competitive results \cite{kim2010gaussian,kandemir2016variational,haussmann2017variational,wu_combining_2021,wang2021multiple,lopez2022deep}, due to their high expressiveness and uncertainty quantification capabilities. 
Most of these GP-based MIL methods build on the popular VGPMIL \cite{haussmann2017variational}, which formulates the MIL problem through sparse GPs for classification using the logistic function. 

In order to achieve mathematical tractability of the logistic function, VGPMIL introduces a variational bound known as the Jaakkola bound. 
Therefore, the training objective becomes a lower bound of the real one. As recently shown in \cite{wang2021multiple}, this theoretical approximation degrades the predictive performance in practice. 
An alternative and exact treatment of the logistic function has recently been introduced in the context of GPs for supervised classification, see \cite{durante2019conditionally,wenzel2019efficient,polson2013bayesian}. 
The idea is to augment the model using Pólya-Gamma variables, obtaining an equivalent and tractable formulation. 
However, to the best of our knowledge, these ideas have never been adapted to the MIL scenario. 

{
\color{review}
In this work, we first reformulate the VGPMIL model using Pólya-Gamma random variables. 
We find that this new model, called PG-VGPMIL, is equivalent to the existing VGPMIL when performing closed-form variational inference updates (they lead to the same update equations).
This phenomenon, which was also observed in supervised classification \cite{durante2019conditionally,wenzel2019efficient}, finds its justification in the properties of the Hyperbolic Secant density \cite{harkness1968generalized}. Thanks to them, the logistic observation model admits two representations \cite{palmer2006variational}, as a Super Gaussian and as a Gaussian Scale Mixture (GSM), that lead to the same variational optimization objective \cite[Theorem 2.1]{durante2019conditionally}.
We build upon the GSM representation to formulate $\psi$-VGPMIL, a general model where $\psi$ is a differentiable GSM density. When $\psi$ is the Hyperbolic Secant density, we recover the original VGPMIL. Although $\psi$-VGPMIL is formulated using GPs, it can be extended to other related probabilistic frameworks, such as Relevance Vector Machines \cite{tipping2001sparse,martino2021joint}.
}

Inspired by the above connection and by the definition of Pólya-Gamma variables \cite{polson2013bayesian}, we replace the Pólya-Gamma distribution with the Gamma distribution, obtaining a new \textcolor{review}{GSM density and thus a new particularization of $\psi$-VGPMIL, which we refer to as G-VGPMIL}. 
The proposed algorithm is evaluated through a comprehensive set of experiments involving different datasets, baselines and metrics (both at the instance and bag level). 
First, we focus on the G-VGPMIL method in a controlled experiment built around the MNIST dataset. 
This allows us to understand its behavior in practice and devise its main properties. Second, we utilize two historically important MIL benchmark datasets (MUSK1 and MUSK2). 
This second experiment again shows that G-VGPMIL improves the existing VGPMIL approach in terms of efficiency and performance. 
Third, we show that G-VGPMIL achieves better results than state-of-the-art methods for the ICH detection problem.

In summary, the main contributions of this work are the following:
\begin{itemize}
    \item We introduce PG-VGPMIL, a new probabilistic MIL model based on Pólya-Gamma random variables. To the best of our knowledge, Pólya-Gamma variables have never been used before in the context of MIL. We observe that PG-VGPMIL is equivalent to the existing VGPMIL when performing closed-form variational inference updates.
    \item \textcolor{review}{Building on the theory behind this equivalence,} we develop \textcolor{review}{$\psi$-VGPMIL, }a general inference framework for the logistic observation model. New inference models can be obtained using different GSM densities, denoted by $\psi$. PG-VGPMIL (and hence VGPMIL) becomes a particular case of this framework \textcolor{review}{when $\psi$ is the Hyperbolic Secant density. }
    \item We use the Gamma distribution to obtain a new GSM density, which is used in the aforementioned general framework to obtain G-VGPMIL, a new inference model for MIL. To the best of our knowledge, the Gamma distribution has never been used in the context of MIL. 
    \item The newly proposed G-VGPMIL is compared to state-of-the-art approaches using different metrics and datasets, including a real-world ICH detection problem. The experiments conducted show enhanced predictive performance and efficiency.
\end{itemize}

The rest of the paper is organized as follows. 
In Section \ref{section:gps4mil} we introduce the MIL problem and its GP-based formulation. 
In Section \ref{section:pg4gpmil} we introduce PG-VGPMIL and \textcolor{review}{investigate} its equivalence to VGPMIL.
In Section \ref{section:general_model} we devise \textcolor{review}{$\psi$-VGPMIL,} a new general framework of which VGPMIL \textcolor{review}{is a} particular realization. 
In Section \ref{section:gamma4gpmil} we introduce our new model G-VGPMIL as another particular realization of $\psi$-VGPMIL. 
In Section \ref{section:results} we empirically validate the newly proposed method. 
In Section \ref{section:conclusion} we summarize the main conclusions of this work.

\section{Gaussian Processes for Multiple Instance Learning}\label{section:gps4mil}

The goal of this section is to introduce the MIL problem and how GPs are used to address it. 

\subsection{Problem formulation: Multiple Instance Learning} \label{subsection:problem}

We consider a dataset $\left\{ (\bx_n, y_n) \colon n \in \left\{ 1, \ldots, N \right\} \right\} \subset \R^D \times \left\{ 0,1 \right\} $, which consists of the training instances $\bx_n$ and their binary labels $y_n$. 
We express this set in a matrix $\bX = \left[ \bx_1, \ldots, \bx_N \right] \in \R^{D \times N}$ and a vector $\by = \left[ y_1, \ldots, y_N \right]^\top \in \left\{ 0,1 \right\}^{N}$.

In the MIL scenario, the instance labels $y_n$ are not observed. 
Instead, they are grouped into bags and only the maximum of the labels of the instances in a bag is observed. 
Formally, the index set $\left\{ 1, \ldots, N \right\}$ is partitioned into $B$ non-overlapping bags $\left\{ \mathrm{Bag}_1, \ldots, \mathrm{Bag}_B \right\}$ and the operator $\left\{ \cdot \right\}_b$ is used to refer to the elements in bag $b$, so $\left\{ \by \right\}_b = \left\{ y_i \colon i \in \mathrm{Bag}_b \right\}$. 
The operator $\left\{ \cdot \right\}_{b \setminus n}$ is used to refer to the elements in bag $b$ except for the n-th element, so $\left\{ \by \right\}_{b \setminus n} = \left\{ \by \right\}_b \setminus \left\{ y_n\right\}$. 
Finally, for a bag $b$, the only observed label is  $T_b =\max \left\{ \by \right\}_b$.
The goal is to predict the label for any new bag, as well as the labels of the instances in the bag.

Notice that the ICH detection problem presented in the introduction can be cast as an MIL one. 
A full scan is treated as a bag and slices are treated as instances. 
Only bag labels $T_b$ are observed, while slice labels $y_n$ are unobserved. 
A CT scan presents hemorrage (positive class) if at least one of its slices show evidence of hemorrhage. A negative scan contains only normal slices.

\subsection{Variational Gaussian Processes for Multiple Instance Learning} \label{subsection:gps4mil}

Before considering the GP MIL problem, we briefly summarize the general framework for GP Logistic classification \cite{rasmussen2006gaussian}.
Given a training dataset $\left( \bX, \by \right)$, GP classification assumes the existence of a latent function $f \colon \R^D \to \R$ that determines the class of an instance.
The model places a GP prior over this function, $\p(\bff) = \mathcal{N}\left(\bff \mid \bzero, \bK_{\bX \bX} \right)$, and uses the observation model
\begin{equation}
    \p(y_n \mid f_n) = \operatorname{Bernouilli}\left( y_n \mid \logit(f_n) \right), \quad \p(\by \mid \bff) = \prod_{n=1}^{N} \p(y_n \mid f_n) \label{eq:y_prob},
\end{equation}
where $\logit(t) = 1/(1+e^{-t})$ is the logistic function and $\bff = \left[ f_1, \ldots,  f_N \right]^\top = \left[ f(\bx_1), \ldots , f(\bx_N) \right]^\top$ are the realizations of the latent function $f$ over the training set. The matrix $\bK_{\bX \bX}$ is defined as $\left(\bK_{\bX \bX} \right)_{i j} = k(\bx_i, \bx_j)$, where $k \colon \R^D \times \R^D \to \R$ is a positive-definite kernel \cite{rasmussen2006gaussian}. \textcolor{review}{In this work, we will use the popular Radial Basis Function (RBF) kernel, which is defined as $\operatorname{k}(\bx, \bx') = v \exp\left(- \frac{\left\| \bx - \bx' \right\|}{2 l} \right)$, where $v, l > 0$ will be treated as hyperparameters. In the following, we denote $\blambda = \left\{v, l\right\}$. }


GP inference requires inverting an $N\times N$ matrix related to the kernel. This has a cost of $O(N^3)$, so the full model can not be used for large datasets. This has motivated the popularization of \emph{sparse GP} methods. Following the \textit{fully independent training conditional (FITC)} approximation \cite{snelson2005sparse}, we introduce a set of $M$ inducing points $\bZ = \left[ \bz_1, \ldots, \bz_M \right]$ and their corresponding output via $f$, $\bu = f(\bZ)$. Mirroring the relation between $\bX$ and $\bff$, we have
\begin{align}
	\p(\bu) & = \mathcal{N}\left( \bu \mid \bzero, \bK_{\bZ \bZ} \right), \label{eq:u_prob}\\
	\p(\bff\mid \bu) & = \mathcal{N}(\bff\mid \ba, \widetilde{\bK}), \label{eq:f_prob}
\end{align}	
where $\ba = \bK_{\bX \bZ} \bK_{\bZ \bZ}^{-1} \bu$ and $\widetilde{\bK} = \bK_{\bX \bX} - \bK_{\bX \bZ} \bK_{\bZ \bZ}^{-1} \bK_{\bZ \bX}$. Here,  $\left(\bK_{\bZ \bZ} \right)_{i j} = k(\bz_i, \bz_j)$, $\left(\bK_{\bX \bZ} \right)_{i j} = k(\bx_i, \bz_j)$ and $\bK_{\bZ \bX} = \bK_{\bX \bZ}^\top$. This approach reduces the cost to $O\left(M^2 N\right)$. 

In the MIL scenario, instead of instance labels $\{y_n\}$ we have only bag labels $T_b$. To adapt the previous classifier to this setting, VGPMIL \cite{haussmann2017variational} introduces the following expression of the bag label likelihood given the labels of the instances,
\begin{equation}\label{eq:bag_prob}
	\p(T_b \mid \left\{ \by \right\}_b) = \frac{H^{G_b}}{H+1}, \quad \p(\bT \mid \by) = \prod_{b=1}^{B}\p(T_b \mid \left\{ \by \right\}_b),
\end{equation}
where $H>0$ (set to 100 in \cite{haussmann2017variational}) and $G_b = T_b \max \left\{ \by \right\}_b + (1 - T_b) \left( 1 - \max \left\{ \by \right\}_b \right)$. This replaces the deterministic value of $T_b$ given $\max\{\by\}_b$ by a noisy probabilistic one. If $\max\{\by\}_b=1$, then $\p(T_b=1 \mid \{\by\}_b)=H/(H+1)$ and $\p(T_b=0 \mid \{\by\}_b)=1/(H+1)$ (a very small number). On the contrary, if $\max\{\by\}_b=0$, then $\p(T_b=0 \mid \{\by\}_b)=H/(H+1)$ and $\p(T_b=1 \mid \{\by\}_b)=1/(H+1)$ (again, a very small number). The complete VGPMIL model is given by the product of the distributions in Eq. \eqref{eq:y_prob}, \eqref{eq:u_prob}, \eqref{eq:f_prob},  and \eqref{eq:bag_prob}. 

In order to make predictions in VGPMIL it is necessary to compute the posterior distribution $\p(\bu, \bff, \by \mid \bT)$, which is not analytically tractable. The original work \cite{haussmann2017variational} resorts to mean-field variational inference \cite{bishop2006pattern} to approximate it by a variational distribution $\q(\bu, \bff, \by) = \q(\bu) \p(\bff \mid \bu) \q(\by)$ with $\q(\by) = \prod_{n=1}^N\q(y_n)$, that is selected minimizing the Kullback-Leibler (KL) divergence between the variational distribution approximation and the true posterior. The solution for each factor is given by \cite[Eq. (10.9)]{bishop2006pattern}. 


Since the logistic function used in Eq. \eqref{eq:y_prob} is not conjugate to the Gaussian distribution, it is impossible to obtain tractable expressions for the variational factors. 
To deal with this, VGPMIL uses the Jaakkola bound \cite{jaakkola2000bayesian}. 
In the next subsection we \textcolor{review}{explore} an augmented version of VGPMIL based on Pólya-Gamma random variables \cite{polson2013bayesian}, \textcolor{review}{in which inference is tractable}.

\section{Pólya-Gamma variables for GPMIL}\label{section:pg4gpmil}

The class of Pólya-Gamma random variables was introduced in \cite{polson2013bayesian} to propose a \textit{data augmentation} approach for stochastic Bayesian inference in models with binomial likelihood.
Later, the multinomial distribution was reformulated in terms of Pólya-Gamma variables considering GPs with multinomial observations \cite{linderman2015dependent}. 
Finally, \cite{wenzel2019efficient} proposed a stochastic variational approach to GP classification using an augmented model based on Pólya-Gamma variables and inducing points. 
Next, we consider the Pólya-Gamma trick in the context of MIL GP classification.

The Pólya-Gamma distribution $\operatorname{PG}\left(b, c\right)$, with $b >0$ and $c \in \R$, has an important property that is closely related to our problem,
\begin{equation}\label{eq:pg-sigmoid}
	\logit(x) = 2^{-1}\exp \left( x/2 \right) \Ebb_{\PG(\omega \mid 1, 0)} \left[ \exp \left(- x^2 \omega / 2  \right) \right], \quad \forall x \in \R.
\end{equation}
This equality ensures that the following joint density is well defined
\begin{equation}\label{eq:pgvgpmil_p_y_omega_}
    \p(y_n, \omega_n \mid f_n) = 2^{-1} \exp\left(( y_n-1/2)f_n\right) \exp\left(- f_n^2 \omega_n / 2 \right) \PG(\omega_n \mid 1, 0),
\end{equation}
and that $\p(y_n \mid f_n) = \int_{0}^{\infty} \p(y_n, \omega_n \mid f_n) \textnormal{d}\omega_n$ is the logistic observation model. 
Writing $\bomega = \left[ \omega_1, \ldots, \omega_N \right]^\top$, $\bOmega = \diag\left( \omega_1, \ldots, \omega_N \right)$, and assuming independence between instances, we have
\begin{align}
    \p(\by, \bomega \mid \bff) = 2^{-N} \exp  \left( (\by - 2^{-1} \mathbf{1})^{\top} \bff - 2^{-1}\bff^{\top} \bOmega \bff \right) \prod_{n=1}^{N}\PG(\omega_n \mid 1, 0), \label{eq:vgpmil_y_omega_prob}
\end{align}
where $\mathbf{1} = \left[ 1, \ldots 1\right]^\top$. Note that we are considering an equivalent model where we have removed the use of the logistic function without introducing any approximation. 
The augmented VGPMIL model is defined by the product of the distributions in Eq. \eqref{eq:u_prob}, \eqref{eq:f_prob}, \eqref{eq:bag_prob} and \eqref{eq:vgpmil_y_omega_prob}. 
We will refer to this model by \textit{Pólya Gamma Variational Gaussian Process Multiple Instance Learning (PG-VGPMIL)}.

\textbf{Inference in PG-VGPMIL.} We mimic the inference procedure followed by VGPMIL. 
We approximate the posterior distribution $\p(\bu, \bff, \by, \bomega \mid \bT)$ with a variational distribution $\q(\bu, \bff, \by, \bomega) = \q(\bu) \p(\bff \mid \bu) \q(\by) \q(\bomega)$, where $\q(\by) = \prod_{n=1}^N\q(y_n)$, and $\q(\bomega) = \prod_{n=1}^N\q(\omega_n)$, minimizing the KL divergence between them. Applying \cite[Eq. (10.9)]{bishop2006pattern}, we obtain
\begin{align}
	\q(\bu) & = \mathcal{N}\left( \bu \mid \bm, \bS \right), \label{eq:pgvgpmil_q_u}\\
	\q(y_n) & = \operatorname{Bernouilli}\left( y_n \mid \pi_n\right), \label{eq:pgvgpmil_q_y}\\
    \q(\omega_n) & = \PG\left( \omega_n \mid 1, c_n \right), \label{eq:pgvgpmil_q_omega}
\end{align}
\vspace{-11mm}
\begin{align}
\hspace{-100mm}
	\pi_n & = \logit \left[ \bK_{\bx_n \bZ}\bK_{\bZ \bZ}^{-1} \bm + \log H \left(2 T_b - 1\right) \left( 1 - \Ebb \left[ \max \left\{ \by \right\}_{ b\setminus n} \right] \right) \right], \label{eq:pgvgpmil_pi_update}\\
	\bm & = \bS\bK_{\bZ \bZ}^{-1} \bK_{\bZ \bX} \left(\bpi - 2^{-1}\mathbf{1} \right), \label{eq:pgvgpmil_m_update}\\
	\bS & = \left( \bK_{\bZ \bZ}^{-1} \bK_{\bZ \bX} \bTheta \bK_{\bX \bZ}\bK_{\bZ \bZ}^{-1} +\bK_{\bZ \bZ}^{-1} \right)^{-1}, \label{eq:pgvgpmil_S_update} \\
    \bTheta & = \diag\left( \theta(c_1), \ldots, \theta(c_n) \right), \quad c_n = \sqrt{\Ebb_{\q(f_n)} [f_n^2]}, \quad \theta(c)= \frac{\tanh \left( c/2 \right)}{2c}\label{eq:pgvgpmil_Theta_update}
\end{align}
where $\q(f_n) = \int_{\R^M}\q(\bu) \p(f_n \mid \bu) \dd \bu$ and $\bpi = \left[ \pi_1, \ldots, \pi_n \right]^\top$. 
It is worth mentioning that $\theta(c_n) = \Ebb_{\q(\omega_n)}\left[ \omega_n\right]$. 
The derivation of these equations can be found in Appendix \ref{appendix:variational_updates}. 
Note that training PG-VGPMIL boils down to iterating over the above equations.
\textcolor{review}{If we compare these updates with the ones from VGPMIL, we find that they are identical. We explore this in the following subsection.}

\subsection{On the equivalence between PG-VGPMIL and VGPMIL}\label{subsection:pg4gpmil_equivalence}

{\color{review}
We observe that the update equations for PG-VGPMIL are equivalent to those of VGPMIL. In both models, $\bm$ and $\pi_n$ have the same expressions, recall  \cite[Eq. (12), Eq. (14)]{haussmann2017variational}. Also, the update for $\xi_n$ in VGPMIL is the same as $c_n$ here. Finally, from $\logit(x) = (\tanh(x/2)+1)/2$, we have that $\bLambda = \bTheta$ and so the covariance matrix $\bS$ is also the same \cite[Eq. (11)]{haussmann2017variational}.

This phenomenon was also observed in the context of supervised classification \cite{durante2019conditionally,wenzel2019efficient}, as a consequence of two different representations of the Hyperbolic Secant density, and generalizes naturally to our setting of MIL. Namely, note that the logistic likelihood can be written as
\begin{equation}\label{eq:obs_model_phi}
    \p(y_n \mid f_n) = \pi \exp\left((y_n-2^{-1})f_n\right) \phi(f_n),
\end{equation}
where $\phi(x) = \left(2\pi \cosh\left(x/2\right) \right)^{-1}$ is the Hyperbolic Secant density proposed in \cite{harkness1968generalized}. 
This density prevents us from calculating variational updates analytically. It admits two representations: as a Super Gaussian (SG), and as a Gaussian Scale Mixture (GSM), the former being a consequence of the latter \cite{palmer2006variational,galy2020automated}. 
The SG representation leads to the Jaakkola bound used in VGPMIL. The GSM representation leads to PG-VGPMIL and is obtained from Eq. \eqref{eq:pg-sigmoid},
\begin{equation}\label{eq:phi_gsm}
    \phi(x) = \int_{0}^{+\infty} \mathcal{N}\left( x \mid 0, \omega^{-1} \right) \widehat{\phi}(\omega) \dd \omega, \quad \forall x \in \R,
\end{equation}
where $\widehat{\phi}(\omega) = (2\pi\omega)^{-1} \operatorname{PG}\left( \omega \mid 1, 0 \right)$.
These representations produce inference schemes that may appear different. However, both approaches lead to optimize the same objective, as demonstrated in \cite[Theorem 2.1]{durante2019conditionally} for the supervised logistic regression model. This result follows straightforwardly for our setting involving MIL and GPs.

Besides being responsible for the equivalence between the two approaches presented so far, the Hyperbolic Secant density $\phi$ plays a fundamental role in the VGPMIL variational updates. 
Recall that the matrix $\bTheta$ in Eq. \eqref{eq:pgvgpmil_Theta_update} is computed using the function $\theta$, which is completely determined by $\phi$, see Appendix \ref{appendix:variational_updates}. As pointed out in previous works, one could try to improve the inference procedure by modifying $\theta$ \cite{babacan2012bayesian}. In the following section, we formalize this idea, and show that it corresponds to replacing $\phi$ by a different GSM density. 
}

\section{A general inference framework for the logistic observation model}\label{section:general_model}

{
\color{review}
The GP logistic observation model can be written as
\begin{equation}
    \p(\bu, \bff, \by) = Z^{-1} \mathcal{N}\left( \bff, \bu \right) \exp\left(  \left(\by - 2^{-1}\mathbf{1}\right)^\top \bff \right) \phi(\bff),
\end{equation}
where $Z = \pi^{-N}$, $\phi(\bff) = \prod_{n=1}^{N} \phi(f_n)$, and $\mathcal{N}\left( \bff, \bu \right) = \mathcal{N}\left( \bff \mid \bu \right) \mathcal{N}\left( \bu \right)$ denotes the joint distribution given by equations \eqref{eq:u_prob} and \eqref{eq:f_prob}. To extend this model, first we consider a differentiable density $\psi \colon \R \to \R$ that admits a GSM representation.  Next, we replace $\phi$ by $\psi$, 
\begin{equation}\label{eq:p_ufy_psi}
    \p(\bu, \bff, \by) = Z^{-1} \mathcal{N}\left( \bff, \bu \right) \exp\left(  (\by - 2^{-1}\mathbf{1})^\top \bff \right) \psi(\bff),
\end{equation}
where $\psi(\bff) = \prod_{n=1}^{N} \psi(f_n)$ and the normalization constant is 
\begin{equation}
    Z = \sum_{\by \in \left\{ 0, 1 \right\}^N } \int_{\R^N} \mathcal{N}\left( \bff \right) \exp\left(  (\by - 2^{-1}\mathbf{1})^\top \bff \right) \psi(\bff) \dd \bff,
\end{equation}
where $\mathcal{N}\left( \bff \right) = \int_{\R^M} \mathcal{N}\left( \bff, \bu \right) \dd \bu$. Using $\sum_{\by \in \left\{ 0, 1 \right\}^N } \exp\left(  (\by - 2^{-1}\mathbf{1})^\top \bff \right) = \pi^{-N} \phi(\bff)^{-1}$, we can write 
\begin{equation}\label{eq:psivgpmil_Z}
    Z = \pi^{-N} \int_{\R^N} \mathcal{N}\left( \bff \right) \psi(\bff) \phi(\bff)^{-1} \dd \bff.
\end{equation}
Given that $\psi$ must be bounded, the above integral must be dominated by $\int_{\R^N} \mathcal{N}\left( \bff \right) \phi(\bff)^{-1} \dd \bff$. As $\mathcal{N}\left( \bff \right)$ is a Gaussian and $\phi(\bff)^{-1}$ is analytic (as a product of analytic functions), then $\mathcal{N}\left( \bff \right) \phi(\bff)^{-1}$ is integrable \cite{zinn2021quantum}. Therefore, the integral in Eq. \eqref{eq:psivgpmil_Z} is finite.

In our new framework, $\p(\by \mid \bff)$ remains the same as in Eq. \eqref{eq:y_prob}, but the GP prior distribution change according to 
\begin{gather}
    \p(\bff \mid \bu) \propto \mathcal{N}\left( \bff, \bu \right) \psi(\bff) \phi(\bff)^{-1}, \label{eq:psi-vgpmil_p_f_u} \\
    \p(\bff) \propto \mathcal{N}\left( \bff \right) \psi(\bff) \phi(\bff)^{-1}.
\end{gather}
This means that we no longer consider a GP prior on $\p(\bff)$, which is now a Gaussian weighted by the ratio $ \psi(\bff) \phi(\bff)^{-1} $. This new approach increases the flexibility of our model, since we can explore different options for $\psi$ and subsequently select the best for our problem. One way to accomplish this is considering a parametric family and looking for the optimal parameters. Recall that the original model can be recovered by setting $\psi=\phi$, which highlights that the new framework is an extension of the previous one. 

To adapt this new framework to MIL, we multiply the joint distribution in Eq. \eqref{eq:p_ufy_psi} by the bag likelihood defined in Eq. \eqref{eq:bag_prob}. We will refer to this extended model as \emph{$\psi$ Variational Gaussian Processes Multiple Instance Learning ($\psi$-VGPMIL)}. 


\textbf{Inference in $\psi$-VGPMIL.} We leverage the GSM representation of $\psi$ and perform inference in the augmented model, similar to PG-VGPMIL. Thus, we approximate the posterior distribution $\p(\bu, \bff, \by \mid \bT)$ with a variational distribution $\q(\bu, \bff, \by) = \q(\bu) \q(\bff \mid \bu) \q(\by)$, where $\q(\bff \mid \bu) = \mathcal{N}(\bff \mid \bu)$, and $\q(\by) = \prod_{n=1}^N\q(y_n)$. 

Our choice of $\q(\bff \mid \bu)$ is not optimal, but it allows us to derive tractable expressions for the variational updates. Note that $\q(\bff \mid \bu) = \p(\bff \mid \bu)$ is a better approximation \cite{titsias:2009}, but yields intractable expressions since $\p(\bff \mid \bu)$ is no longer a Gaussian, see Eq. \eqref{eq:psi-vgpmil_p_f_u}. Our approach could be improved by giving more flexibility to $\q(\bff \mid \bu)$, e.g. using normalizing flows \cite{kobyzev2020normalizing}.

The optimal expressions for $\q(\bu)$ and $\q(\by)$ are obtained minimizing the KL divergence between the posterior and the variational distributions. Applying \cite[Eq. (10.9)]{bishop2006pattern}, we obtain
\begin{align}
	\q(\bu) & = \mathcal{N}\left( \bu \mid \bm, \bS \right), \label{eq:psi-vgpmil_q_u}\\
	\q(y_n) & = \operatorname{Bernouilli}\left( y_n \mid \pi_n\right), \label{eq:psi-vgpmil_q_y}
\end{align}
\vspace{-11mm}
\begin{align}
	\pi_n & = \logit \left[ \bK_{\bx_n \bZ}\bK_{\bZ \bZ}^{-1} \bm + \log H \left(2 T_b - 1\right) \left( 1 - \Ebb \left[ \max \left\{ \by \right\}_{ b\setminus n} \right] \right) \right], \label{eq:psi-vgpmil_pi_update}\\
	\bm & = \bS\bK_{\bZ \bZ}^{-1} \bK_{\bZ \bX} \left(\bpi - 2^{-1}\mathbf{1} \right), \label{eq:psi-vgpmil_m_update}\\
	\bS & = \left( \bK_{\bZ \bZ}^{-1} \bK_{\bZ \bX} \bTheta \bK_{\bX \bZ}\bK_{\bZ \bZ}^{-1} +\bK_{\bZ \bZ}^{-1} \right)^{-1}, \label{eq:psi-vgpmil_S_update} \\
    \bTheta & = \diag\left( \theta(c_1), \ldots, \theta(c_N) \right), \quad c_n = \sqrt{\Ebb_{\q(f_n)} [f_n^2]}, \quad \theta(c) = -\frac{\psi'(c)}{c\psi(c)}, \label{eq:psi-vgpmil_Theta_update}
\end{align}
where $\q(f_n) = \int_{\R^M}\q(\bu) \p(f_n \mid \bu) \dd \bu$ and $\bpi = \left[ \pi_1, \ldots, \pi_n \right]^\top$. The expression of $\theta(c_n)$ reveals why $\psi$ must be differentiable. As in PG-VGPMIL, $\theta(c_n)$ represents the expectation of the augmenting variables. We provide more details on the derivation of these equations in Appendix \ref{appendix:variational_updates}. Again, note that if $\psi = \phi$, we recover the original VGPMIL model

\textbf{Kernel hyperparameters estimation in $\psi$-VGPMIL.} The variational framework allows us to estimate the kernel hyperparameters $\blambda$. Following \cite{hensman2015mcmc}, we aim to maximize the ELBO with respect to them. This is computationally equivalent to placing a flat improper prior $\p(\blambda) \propto \operatorname{const}$, and taking the mode of the approximated posterior $\q(\blambda)$ as an estimate. The objective to be maximized is
\begin{align}\label{eq:psi-vgpmil_hyperparams_obj}
    \mathcal{J}(\blambda) = & - \operatorname{KL}\left(\q(\bu), \p(\bu)\right) + \left( \bpi - 2^{-1} \bone \right)^\top \bmu + \nonumber \\
    & + \sum_{n=1}^{N} \mathbb{E}_{\q(f_n)} \left[ \log \psi(f_n) \right] - \log Z,
\end{align}
where $\bmu = \bK_{\bX \bZ} \bK_{\bZ \bZ}^{-1} \bm$ is the mean of $\q(\bff)$. 
We optimize $\mathcal{J}(\blambda)$ using gradient ascent for a fixed number of iterations, for which we approximate the expectation on the third and fourth terms using Monte Carlo sampling \cite{mohamed2020monte}. 
Note that $Z$ depends on the kernel hyperparameters through $\mathcal{N}\left( \bff\right)$. Since $\log Z$ requires sampling from $\mathcal{N}(\bff)$, which is costly, we decide to further approximate those samples using Random Fourier Features \cite{hensman2017variational}.

It is worth mentioning that the hyperparameter estimation procedure was not implemented in the original VGPMIL. Thus, our work strictly generalizes the model proposed in \cite{haussmann2017variational}. 

The training procedure of $\psi$-VGPMIL is detailed in Algorithm \ref{alg:psi-vgpmil}. An iteration updates the variational parameters using Eq. \eqref{eq:psi-vgpmil_pi_update} - \eqref{eq:psi-vgpmil_Theta_update}, and then updates the kernel hyperparameters optimizing the objective in Eq. \eqref{eq:psi-vgpmil_hyperparams_obj}.

\begin{algorithm}
\footnotesize
\caption{Training procedure of \textcolor{review}{$\psi$-VGPMIL}.}\label{alg:psi-vgpmil}
\KwInput{Training instances $\bX = \left\{ \bx_1, \ldots, \bx_N \right\}$, \textcolor{review}{bag indices $\left\{ \mathrm{Bag}_1, \ldots, \mathrm{Bag}_B \right\}$,} bag labels $\bT = \left\{ T_1, \ldots, T_B \right\}$, number of inducing points, number of iterations $K$, \textcolor{review}{kernel hyperparameters $\blambda$, function $\psi$.}}
Initialize the locations of the inducing points $\bZ$. 
Compute the kernel matrices $\bK_{\bX \bX}, \bK_{\bX \bZ}, \bK_{\bZ \bX}, \bK_{\bZ \bZ}, \bK_{\bZ \bZ}^{-1}, \widetilde{\bK}$ \textcolor{review}{using $\blambda$}.

\textcolor{review}{Initialize the components of $\bm$ and $\bS$ and to random values drawn from $\mathcal{N}(0,1)$. Initialize the components of $\bpi$ to random values drawn from $\operatorname{Uniform}(0,1)$.}

\For{$k = 1, \ldots, K$}
{
    Update $\bTheta$ using Eq. \eqref{eq:psi-vgpmil_Theta_update}.
    
    Update $\bS$ using Eq. \eqref{eq:psi-vgpmil_S_update}.

    Update $\bm$ using Eq. \eqref{eq:psi-vgpmil_m_update}.
    
    For each $n \in \left\{ 1, \ldots , N\right\}$, update $\pi_n$ using Eq. \eqref{eq:psi-vgpmil_pi_update}. 

    \textcolor{review}{Estimate $\widehat{\blambda} = \arg \max_{\blambda} \mathcal{J}(\blambda)$, with $\mathcal{J}(\blambda)$ given in Eq. \eqref{eq:psi-vgpmil_hyperparams_obj}. Recompute the kernel matrices using $\widehat{\blambda}$.}
    
}
\KwOutput{\textcolor{review}{Kernel hyperparameters $\widehat{\blambda}$,} distributions $\q(\bu)$ and $\q(\by)=\prod_{n=1}^{N} \q(y_n)$ as in Eq. \eqref{eq:psi-vgpmil_q_u} and \eqref{eq:psi-vgpmil_q_y}.}
\end{algorithm}
}

\textbf{Making predictions in $\psi$-VGPMIL.} Given a new bag $\left\{ \bx^{*}_{1}, \ldots , \bx^{*}_{N_*} \right\}$, we are interested in both instance and bag level predictions, which are denoted by $\by^* = \left\{ y^{*}_{1}, \ldots, y^{*}_{N_*} \right\}$ and $T^*$, respectively. 
To predict the latent function value $f^*$ of an instance $\bx^*$, we substitute the approximate posterior into the predictive distribution, 
\begin{equation}\label{eq:prediction_gp}
	\p(f^* \mid \bT) \approx \mathcal{N}\left( f^* \mid \mu^*, {\sigma^*}^2 \right),
\end{equation}
where $\mu^* = \bK_{\bx^* \bZ} \bK_{\bZ \bZ}^{-1} \bm$ and ${\sigma^*}^2 = \bK_{\bx^* \bx^*} + \bK_{\bx^* \bZ} \bK_{\bZ \bZ}^{-1} \left( \bS \bK_{\bZ \bZ}^{-1} - \bI \right) \bK_{\bZ \bx^*}$. To compute the distribution of the test label $y^*$, we integrate with respect to $f^*$ in $\p(y^*, f^* \mid \bT)$ to obtain
\begin{equation}
    \p(y^* = 1 \mid \bT) = \int_{-\infty}^{+\infty} \p(f^* \mid \bT) \logit(f^*) \mathrm{d}f^* = \mathbb{E}_{\p(f^* \mid \bT)} \left[ \operatorname{logit}(f^*) \right],
\end{equation}
which allows us to make instance-level predictions having trained the model exclusively with bag labels. 
{\color{review}
As the expectation cannot be calculated in closed form, we approximate it by sampling from the GP predictive distribution,
\begin{gather}
    \p(y^* = 1 \mid \bT) \approx \widehat{\p} = \frac{1}{L} \sum_{l=1}^{L} \operatorname{logit}(f_{l}^*), \label{eq:mean_instance}\\ 
    \operatorname{Var}_{\p(f^* \mid \bT)}\left[ \operatorname{logit}(f^*) \right] \approx \frac{1}{L} \sum_{l=1}^{L} \left(\operatorname{logit}(f_{l}^*) - \widehat{\p} \right)^2, \label{eq:var_instance}
\end{gather}
where $\left\{ f_{1}^* , \ldots, f_{L}^* \right\} \sim \p(f^* \mid \bT)$. 
}
To obtain the bag label, we apply the MIL hypothesis,
\begin{align}
    \p(T^* = 1 \mid \bT) & = 1 - \prod_{n=1}^{N_*} \p(y^*_n = 0 \mid \bT) = \\
    & = \mathbb{E}_{\nu(\bff^* \mid \bT)}\left[ 1 - \prod_{n=1}^{N_*}\left( 1 - \operatorname{logit}(f_n^*) \right) \right],
\end{align}
where $\nu(\bff^* \mid \bT) = \prod_{n=1}^{N_*} \p(f_n^* \mid \bT)$. 
{\color{review}
Again, we estimate the above expectation using samples from the GP predictive distribution, 
\begin{gather}
    \p(T^* = 1 \mid \bT) \approx \widetilde{\p} =  \frac{1}{L} \sum_{l=1}^L \left( 1 - \prod_{n=1}^{N_*} \left( 1 - \operatorname{logit}(f_{nl}^*)  \right) \right), \label{eq:mean_bag}\\
    \operatorname{Var}_{\nu(\bff^* \mid \bT)}\left[ 1 - \prod_{n=1}^{N_*}\left( 1 - \operatorname{logit}(f_{n}^*) \right) \right] \approx   \frac{1}{L} \sum_{l=1}^L \left( \widetilde{\p} - \left( 1 - \prod_{n=1}^{N_*}\left( 1 - \operatorname{logit}(f_{nl}^*) \right) \right)  \right)^2, \label{eq:var_bag}
\end{gather}
where $\left\{ f_{n1}^* , \ldots, f_{nL}^* \right\} \sim \p(f_n^* \mid \bT)$ for each $n \in \left\{ 1, \ldots, N_*\right\}$.
}

\section{Gamma variables for GP-MIL}\label{section:gamma4gpmil}

As explained in the previous section, new models that generalize VGPMIL can be obtained by replacing the Hyperbolic Secant density with a different density that also admits a GSM representation.  
In this section, we focus on a concrete realization of $\psi$-VGPMIL, which uses Gamma variables and will prove to work better in practice. 

\begin{figure}[htbp]
    \centering
    \subfloat{
        \includegraphics[trim={0cm 0cm 0cm 0cm},clip,width=0.4\textwidth]{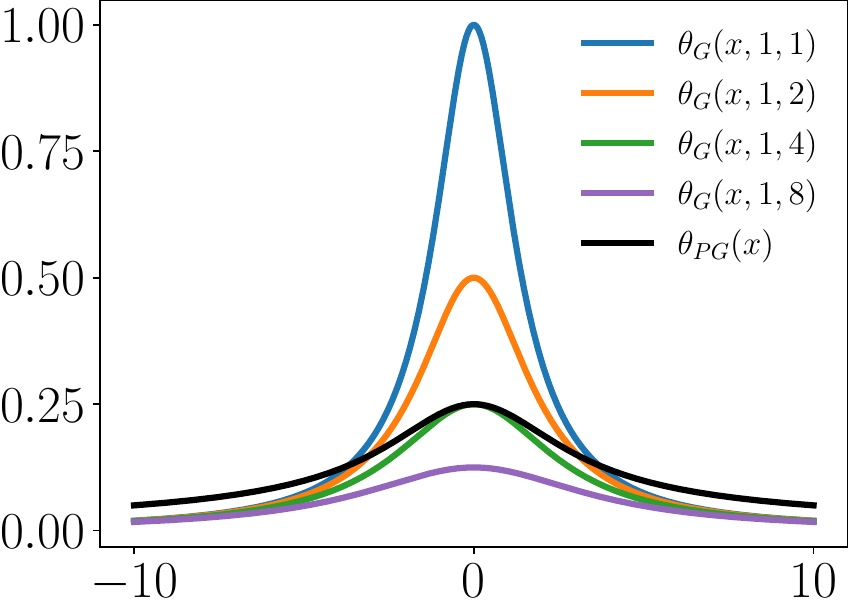}
        }
    \caption{ Plot of the functions $\theta_{\textnormal{PG}}(x) = \tan(x/2)/(2x)$ (black) and $\theta_{\textnormal{G}}(x; \alpha, \beta) = \alpha/(\beta + x^2/2)$ for $\alpha = 1$ and $\beta \in \left\{ 1, 2, 4, 8 \right\}$ (blue, orange, green, purple).}
    \label{fig:theta_functions_plot}
\end{figure}

Our motivation to use Gamma variables is as follows. A Pólya-Gamma variable $\omega \sim \operatorname{PG}(1,0)$ is defined as an infinite weighted sum of independent $\operatorname{Gamma}$ variables \cite[Definition 1]{polson2013bayesian}, \textcolor{review}{$\omega = (2\pi^2)^{-1} \sum_{m=1}^{\infty} g_m/ (m-0.5)^2$}, 
where $g_m \sim \operatorname{Gamma}(1,1)$. We alternatively consider $g_m=g \sim \operatorname{Gamma}(1,1)$ for every $m$, which leads to $\omega = g/4$ and, therefore, $\omega \sim \operatorname{Gamma}(1, 4)$. 
{\color{review}
Thus, we replace the Pólya-Gamma density in Eq. \eqref{eq:phi_gsm} by a Gamma density with parameters $\alpha$ and $\beta$, obtaining 
\begin{equation}\label{eq:psi_gamma}
    \psi(x) = Z(\alpha, \beta)^{-1} \left( \beta + x^2/2 \right)^{-\alpha}, \quad \forall x > 0,
\end{equation}
where $Z(\alpha, \beta)$ is the normalization constant. It is worth mentioning that although this constant can be calculated, it is not needed to carry out the updates in Algorithm \ref{alg:psi-vgpmil}. Thus, we arrive at a concrete realization of $\psi$-VGPMIL, which we call \emph{Gamma Variational Gaussian Processes Multiple Instance Learning (G-VGPMIL)}. 

We note that G-VGPMIL differs from PG-VGPMIL in the function $\theta$ used to compute the $\bTheta$ matrix. In PG-VGPMIL this function is defined as $\theta_{\mathrm{PG}}(x) = \tanh(x/2)/(2x)$, while in G-VGPMIL we obtain $\theta_{\mathrm{G}}(x; \alpha, \beta) = \alpha/(\beta + x^2/2)$. Both functions are plotted in Figure \ref{fig:theta_functions_plot} for $\alpha=1$ and different values of $\beta$. We leave the estimation of $\alpha$ and $\beta$ for future work. 
}

\section{Results}\label{section:results}

In this section, we carry out an empirical validation of the newly formulated G-VGPMIL model by means of three different experiments. Table \ref{table:datasets_info} summarizes the five datasets that we use. First, using an MIL version of the well-known MNIST dataset, we perform a controlled and visual experiment to understand the behavior of our method at both the instance and bag levels. Second, we employ two classical benchmark datasets for MIL algorithms, the MUSK1 and MUSK2 datasets. Third, we tackle the real-world medical problem of ICH detection, showing enhanced performance against state-of-the-art approaches. \textcolor{review}{We compare VGPMIL (Algorithm \ref{alg:psi-vgpmil} with $\psi=\phi$) and G-VGPMIL (Algorithm \ref{alg:psi-vgpmil} with $\psi$ as in Eq. \eqref{eq:psi_gamma}). To ensure a fair comparison, both models were trained with identical set-ups, initial parameters, and grid-searches in every experiment. Both models have been implemented in \texttt{JAX}\footnote{\url{https://jax.readthedocs.io/en/latest/}}, and will be available at \url{https://github.com/Franblueee/psi-VGPMIL}} upon the acceptance of the paper.

\begin{table}[htbp]
	\centering
	\resizebox{\textwidth}{!}{%
		\begin{tabular}{ccccccc}
			\hline
			& Num. instances & Positive instances & Negative instances & Num. bags & Positive bags & Negative bags \\ \hline
			MNIST & 70000 & 8774 & 61226 & 7000 & 3500 & 3500 \\
			MUSK1 & 476 & N/A                    & N/A                    & 92 & 47               & 45               \\
			MUSK2 & 6598 & N/A                    & N/A                    & 102 & 39               & 63               \\ 
			RSNA &  39750 & 5782                    & 33968                   & 1150 &   483             & 667               \\ 
			CQ500 & 193317 & N/A                    & N/A                    & 490 & 205               & 285               \\ \hline
		\end{tabular}
	}
	\caption{ Label distribution at the instance and bag level for each dataset considered.}
	\label{table:datasets_info}
\end{table}

\subsection{MNIST}

\textbf{Overview}.
In this section we analyze the behavior of the newly proposed G-VGPMIL in a controlled environment. To this end, we transform the MNIST dataset \cite{deng2012mnist} into an MIL one. This experiment allows us to understand the performance of our method at both instance and bag level. The results show that G-VGPMIL performs, at least, as well as the state-of-the-art VGPMIL \cite{haussmann2017variational} while taking less time to complete the training. 

\begin{figure}[htbp]
	\centering
	\subfloat[Positive bag with one positive instance.]{
			\fbox{\includegraphics[trim={0cm 0cm 0cm 0cm},clip,width=0.9\textwidth]{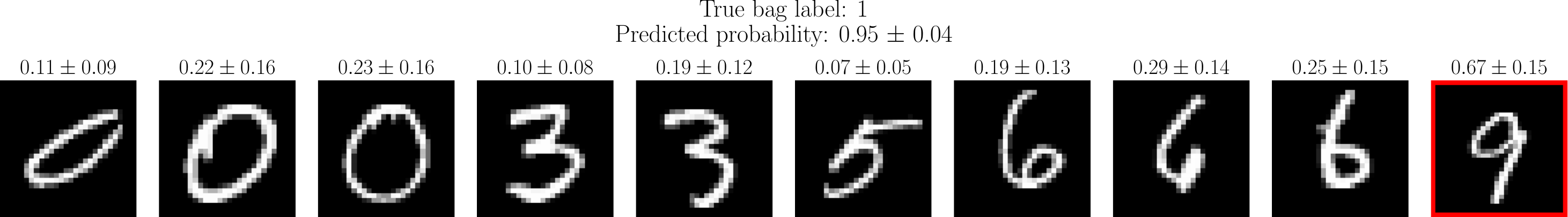}}
	}
    \vspace{0.2cm}
    \subfloat[Positive bag with four positive instances.]{
			\fbox{\includegraphics[trim={0cm 0cm 0cm 0cm},clip,width=0.9\textwidth]{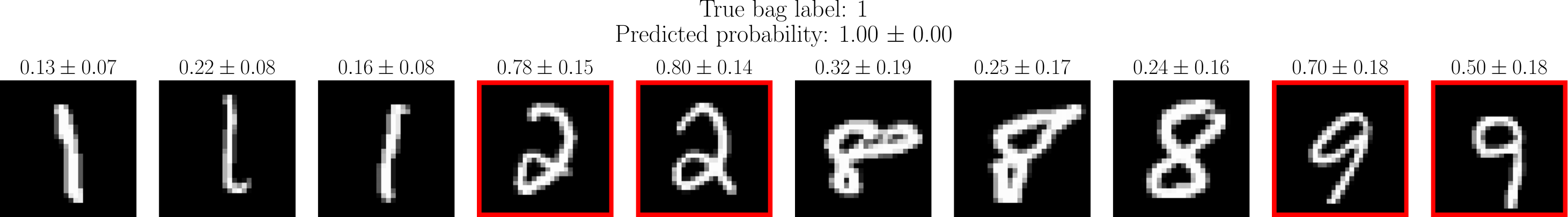}}
	}
	\caption{MNIST bags and G-VGPMIL predictions (probability to be positive). Positive instances are highlighted with a red frame.}
	\label{fig:mnist_bag_prediction_example}
\end{figure}

\textbf{Dataset description}.
The MNIST dataset consists of 70000 images (60000 for training and 10000 for testing) of handwritten digits. \textcolor{review}{We choose the digits 2 and 9 to be the positive class}, while the rest of the digits belong to the negative class. This way, a bag is positive if it contains at least a 2 or a 9, and negative otherwise. We randomly group the digits into bags of \textcolor{review}{10} instances each, \textcolor{review}{ensuring that there is a balanced distribution of positive and negative bags, with each positive bag containing 1 to 4 positive instances}. The resulting dataset has 61226 positive instances, 8774 negative instances and 7000 bags. Figure \ref{fig:mnist_bag_prediction_example} shows two of the generated bags, as well as the predictions obtained by G-VGPMIL. \textcolor{review}{We also show the standard deviation for each prediction, which was calculated from equations \eqref{eq:var_instance} and \eqref{eq:var_bag}.}

We consider two versions of this dataset. In the first one, called MNIST RAW, we use all the $28 \times 28 = 784$ features of the instances without any kind of preprocessing. Given that shallow models may perform poorly when dealing with high-dimensional data, in the second version, i.e., MNIST PCA, we apply Principal Component Analysis (PCA) and retain the first 30 principal components of each instance.

\textbf{Experimental details}. 
For both VGPMIL and G-VGPMIL models, we fix $H=100$ and choose the RBF kernel. \textcolor{review}{The initial kernel hyperparameters are $(v,l)=(0.5, 30)$ for MNIST PCA and $(v,l)=(0.5, 784)$ for MNIST RAW}. We conduct a grid search considering $\left\{50, 100, 200\right\}$ for the number of inducing points, \textcolor{review}{$\left\{ 0.5, 1.0 \right\}$ for the $\alpha$ hyperparameter and $\left\{ 1.0, 2.5, 4.0 \right\}$ for the $\beta$ hyperparameter}. We create five different train-test stratified splits in which we evaluate all previous configurations. To guide training, we monitor the Bag AUC score in a validation subset and halt the process if no improvement occurs during \textcolor{review}{ten} epochs (an epoch is a complete update of the variational parameters). Based on the Bag Accuracy score, we select the best model and report the corresponding cross-validation metrics.

\begin{figure}[htbp]
	\centering
    \subfloat{
		\includegraphics[trim={0cm 0cm 0cm 0cm},clip,width=0.6\textwidth]{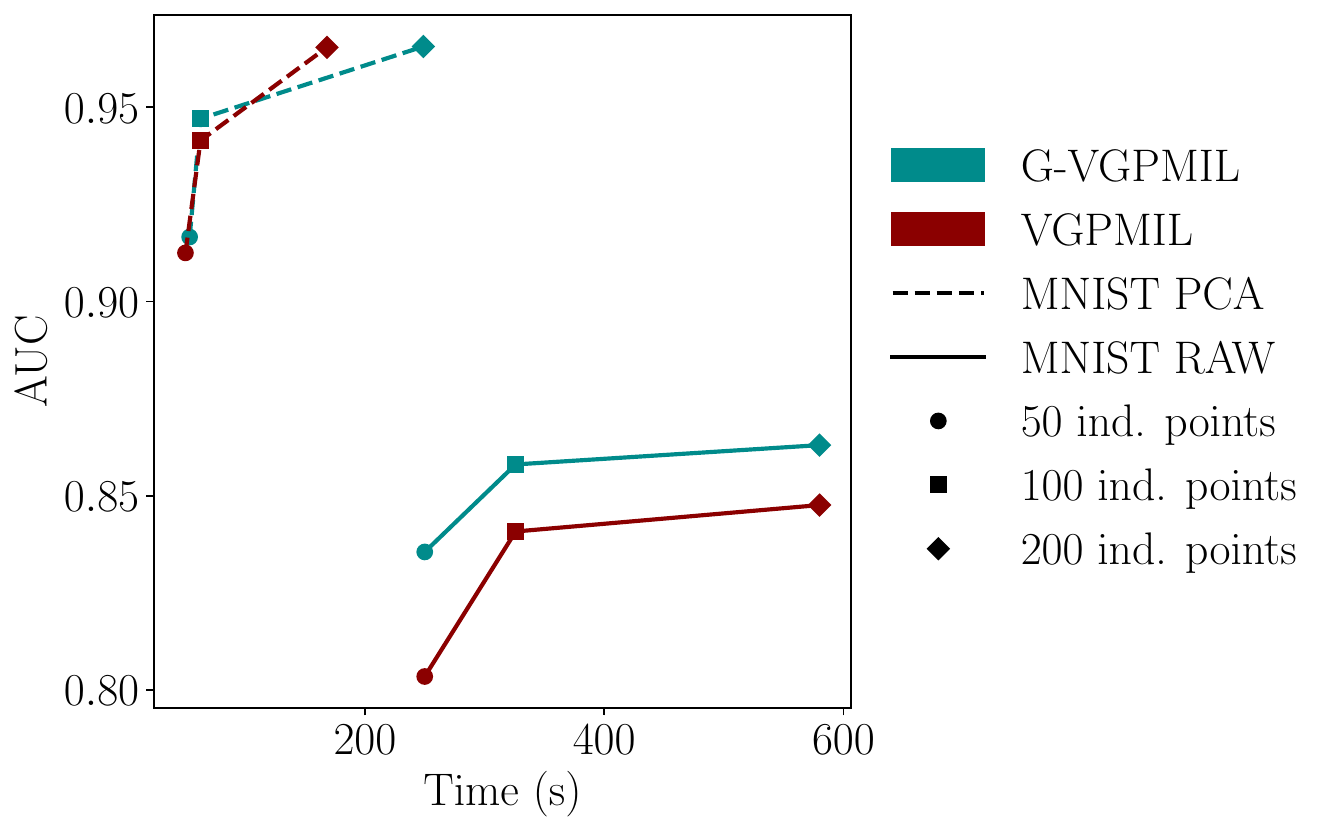}
	}
	\caption{ Training time vs AUC in the MNIST dataset. 
    }
	\label{fig:timevsauc-mnist}
\end{figure}

\textbf{Result 1: \textcolor{review}{competitive performance and training time}}.
In Figure \ref{fig:timevsauc-mnist} we have selected, for each number of inducing points, the best model according to the \textcolor{review}{Bag Accuracy} score, and plotted it against the training time in seconds. 
\textcolor{review}{VGPMIL and G-VGPMIL exhibit similar traning times. They perform comparably in MNIST PCA, while G-VGPMIL is superior in MNIST RAW, suggesting enhanced performance in handling high-dimensional data. Overall, applying PCA improves the performance as it eliminates the existing redundancy in MNIST}. Notice that, as theoretically expected, both the training time and the model performance increase with the number of inducing points.


\begin{figure}[htbp]
	\centering
	\subfloat[\centering  MNIST RAW.]{
            \includegraphics[trim={0cm 0cm 0cm 0cm},clip,width=0.4\textwidth]{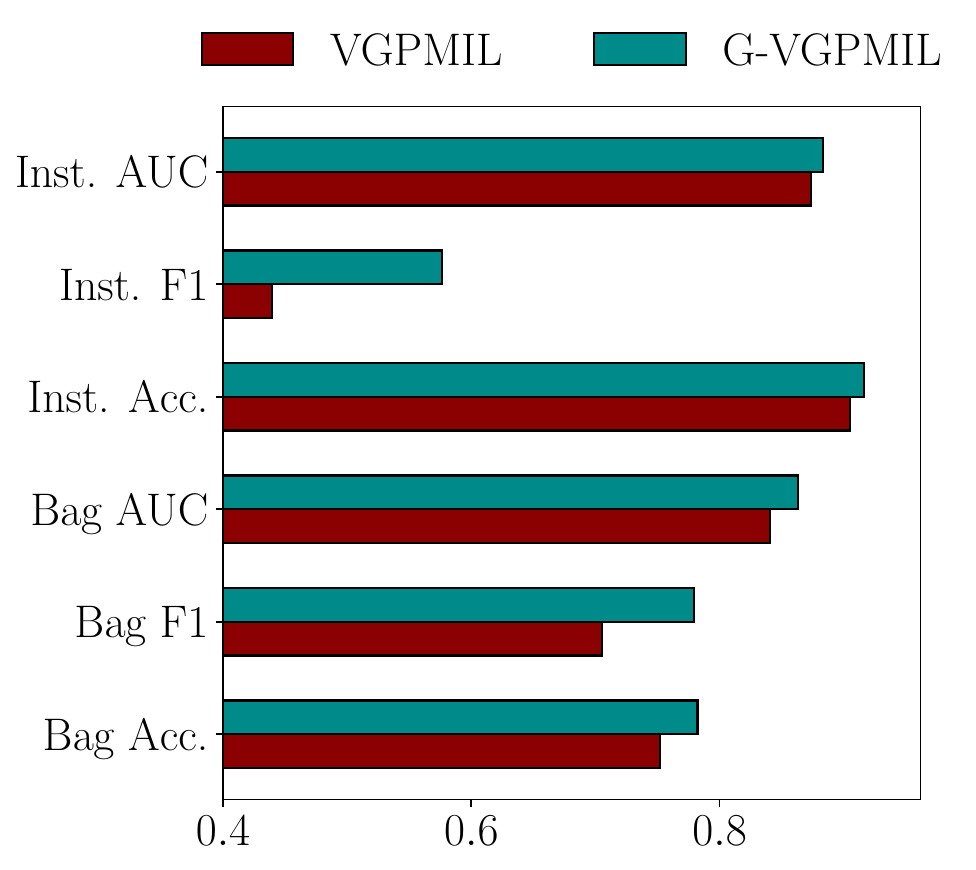}
		}
	\subfloat[\centering MNIST PCA.]{
            \includegraphics[trim={0cm 0cm 0cm 0cm},clip,width=0.4\textwidth]{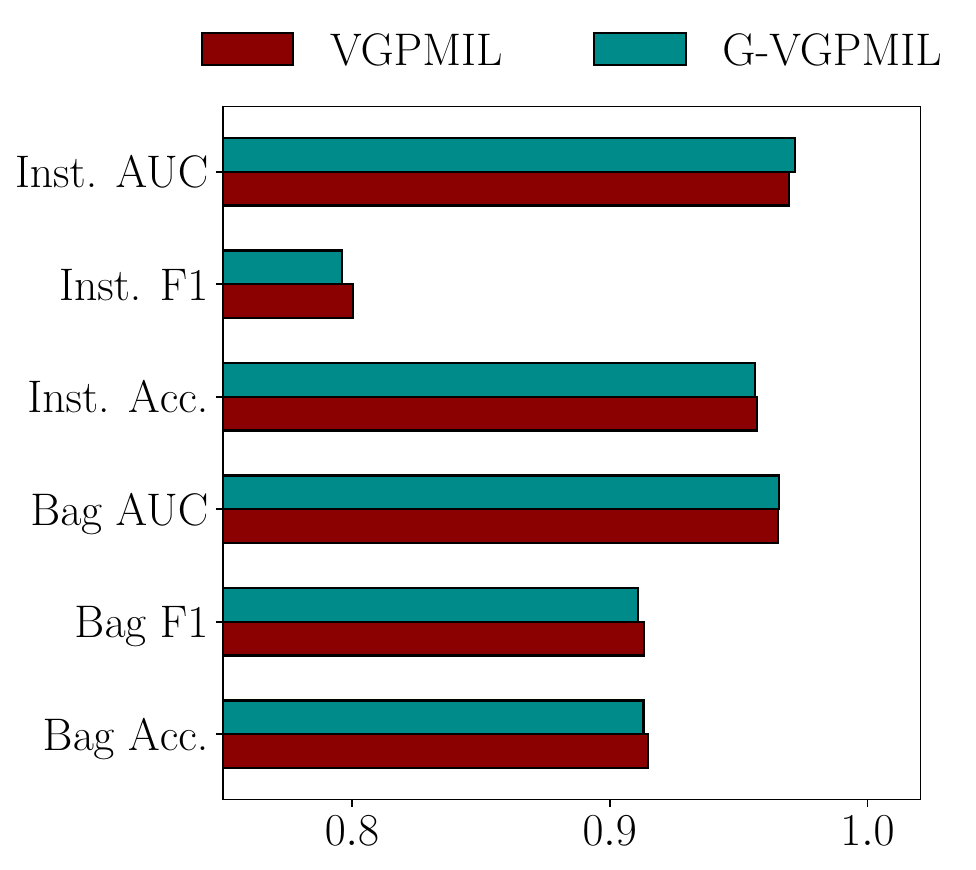}
		}
	\caption{ Bag level and instance level performance in MNIST.}
 \label{fig:bargraph_mnist_allmetrics_best}
\end{figure}

\textbf{Result 2: instance level results and other metrics}. 
In the previous result, we focused on the Bag AUC metric. 
Now, we analyze the AUC, Accuracy and F1-score performance at bag and instance levels.
This is especially important given the high imbalance of the instance labels (61226 negatives vs 8774 positives, recall Table \ref{table:datasets_info}). In Figure \ref{fig:bargraph_mnist_allmetrics_best} we collect the metrics corresponding to the best model in each scenario. These metrics are also available in tables \ref{table:mnist_pca} and \ref{table:mnist_raw}.
In MNIST PCA there are \textcolor{review}{few} differences between VGPMIL and G-VGPMIL, while  in MNIST RAW the differences are \textcolor{review}{significantly} higher in favor of G-VGPMIL, at both the instance and bag level. 


\subsection{MUSK (1 and 2)}

\textbf{Overview.}
In this section, we evaluate G-VGPMIL in two historically important MIL benchmark datasets. Notice that these datasets come from a real problem (see next paragraph). Yet, we show that G-VGPMIL performs similarly to the previous experiment, achieving once again better performance in a similar training time as VPGMIL. 

\textbf{Dataset description.}
MUSK1 and MUSK2 \cite{dietterich1997solving} are two tabular datasets that belong to the domain of drug activity prediction. The goal of this problem is to predict if a molecule will bind to a target binding site (a cavity into which the molecule fits). This is determined by the shapes that the molecule can adopt by rotating its bonds. If at least one of these shapes actually binds to the binding site, the molecule is labeled as positive. Otherwise, it is labeled as negative. This way, a bag represents a whole molecule, while an instance is each of the shapes that the molecule can adopt. MUSK1 has 476 instances and 92 bags (47 positive and 45 negative). MUSK2 has 6598 instances and 102 bags (39 positive and 63 negative). Instances are represented by a 166-dimensional real vector and their labels are unknown, so we will only analyze the metrics at the bag level.  

\textbf{Experimental details.} Again, we fix $H=100$ and choose the RBF kernel. \textcolor{review}{The initial kernel hyperparameters are $(v,l)=(0.5, 166)$}. We perform a grid search considering $\left\{50, 100, 200\right\}$ for the number of inducing points, \textcolor{review}{$\left\{ 0.5, 1.0 \right\}$ for the $\alpha$ hyperparameter and $\left\{ 1.0, 2.5, 4.0 \right\}$ for the $\beta$ hyperparameter}. We create five different train-test stratified splits in which we evaluate each of the previous configurations. To guide the training process we monitor the AUC metric in a validation subset and halt the training when it has not improved for \textcolor{review}{ten} epochs (an epoch is a complete update of the variational parameters). Based on the Bag Accuracy score, we select the best model and report the corresponding cross-validation metrics. 

\begin{figure}
	\centering
	\subfloat{
		\includegraphics[trim={0cm 0cm 0cm 0cm},clip,width=0.6\textwidth]{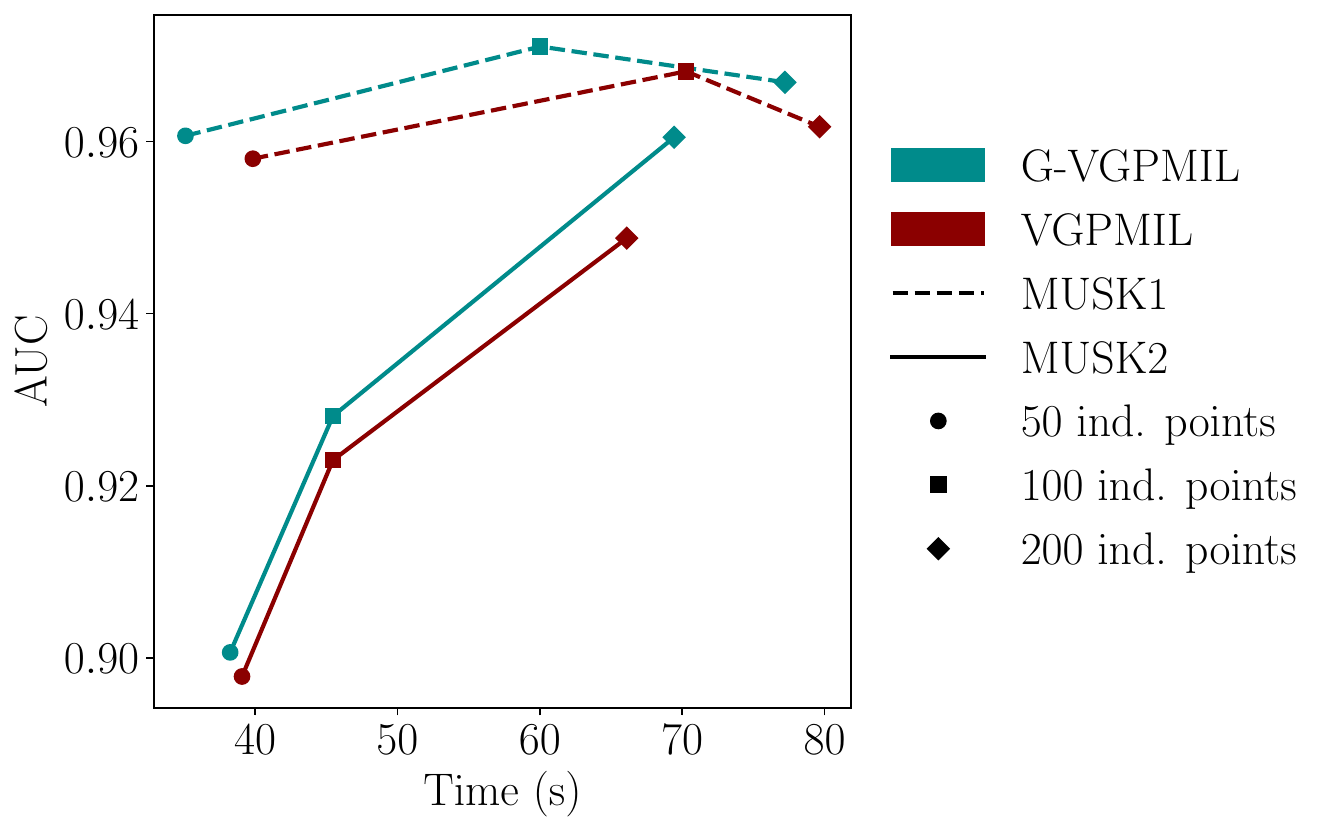}
	}
	\caption{ Training time vs AUC for MUSK1 and MUSK2. 
    }
	\label{fig:timevsauc-musk}
\end{figure}

\textbf{Result 1: better performance in similar training time.} In Figure \ref{fig:timevsauc-musk} we show a plot analogous to that of the MNIST experiment. We have selected the best model (according to the Bag AUC metric) for each number of inducing points considered, and have plotted it against the training time in seconds. For easier comparison, MUKS1 and MUSK2 results are shown together. Observe that the blue dots (G-VGPMIL) are above the red ones (VGPMIL), which means that our method shows superior performance. \textcolor{review}{In terms of the training time, G-VGPMIL is clearly faster in MUSK1, while it is competitive with VGPMIL in MUKS2.}


\begin{figure}
	\centering
	\subfloat[\centering MUSK1.]{
        \includegraphics[trim={0cm 0cm 0cm 0cm},clip,width=0.4\textwidth]{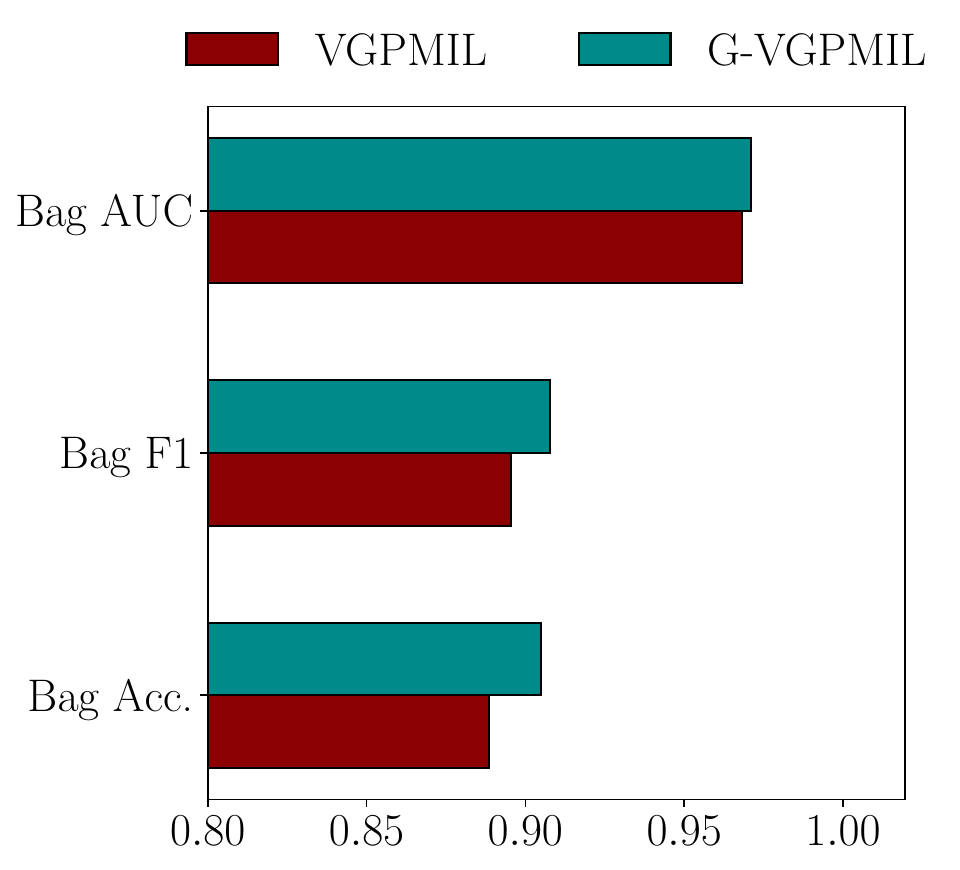}
	}
	\subfloat[\centering MUSK2.]{
        \includegraphics[trim={0cm 0cm 0cm 0cm},clip,width=0.4\textwidth]{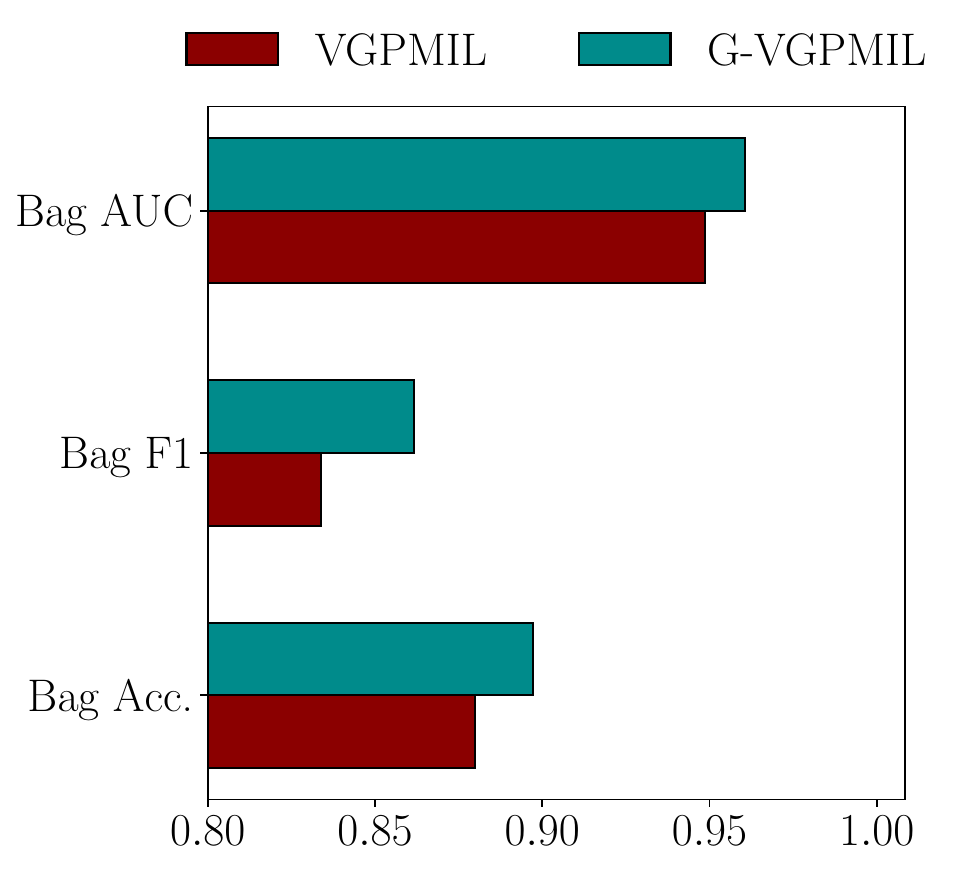}
	}
	\caption{ Bag level metrics for MUSK1 and MUSK2.}
        \label{fig:bargraph_musk_allmetrics_best}
\end{figure}

\textbf{Result 2: other bag level metrics.} Remember that in the MUSK1 and MUSK2 datasets we do not have the instance level labels (recall Table \ref{table:datasets_info}), so we cannot assess the performance at this level. However, we analyze other bag level metrics to ensure that our method performs correctly in all aspects. For MUSK2 the F1 metric is very important given the imbalance between positive bags and negative bags (see Table \ref{table:datasets_info}). Figure \ref{fig:bargraph_musk_allmetrics_best} shows the metrics corresponding to the best model for each of the datasets (also available in tables \ref{table:musk1} and \ref{table:musk2}). Observe that G-VGPMIL attains better values than VGPMIL, \textcolor{review}{especially in MUSK2}. The superiority of G-VGPMIL are consistent with the behavior we observed in MNIST: our approach performs better when dealing with high-dimensional data, which is exactly the case here.

\subsection{Intracranial hemorrage detection (RSNA, CQ500)}

\textbf{Overview.} So far, we have focused on understanding the behavior of G-VGPMIL (we have compared only with VGPMIL) and we have worked with two relatively easy problems. 

Here, we consider a more complex real-world medical problem of detecting ICH from brain CT scans, for which we need to perform a more complex feature extraction process, which will be explained below.
Moreover, we will provide a wider comparison against various state-of-the-art approaches that have been used in this same task. In summary, we will see that G-VGPMIL achieves very good results in predictive performance, efficiency and stability.

\textbf{Dataset description.} We will use two datasets to train and evaluate our method. The first one was published by the Radiological Society of North America (RSNA)\footnote{\url{https://www.kaggle.com/competitions/rsna-intracranial-hemorrhage-detection/}} in 2019. It includes a total of 39750 slices from 1150 patients. The slice (instance) labels are known: there are 5782 abnormal slices (positive) and 33968 normal slices (negative). Regarding the scans (bags), there are 483 abnormal scans (positive) and 667 normal scans. The slices are of size $512 \times 512$ and the number of them in each scan varies from 24 to 57. The results for this dataset are shown in Subsection \ref{subsection:results_ich_rsna}. We will also use the CQ500 dataset \cite{chilamkurthy2018deep}, which was acquired from different institutions in New Delphi, India, as an independent test set. It includes 193317 slices and 491 bags, with only labels at bag level (205 abnormal and 286 normal). The number of slices in each scan varies from 16 to 128. The results for this dataset are shown in Subsection \ref{subsection:results_ich_cq500}.

\begin{figure}[htbp]
	\centering
	\subfloat{
		\includegraphics[trim={2cm 6.3cm 1.8cm 3cm},clip,width=0.8\textwidth]{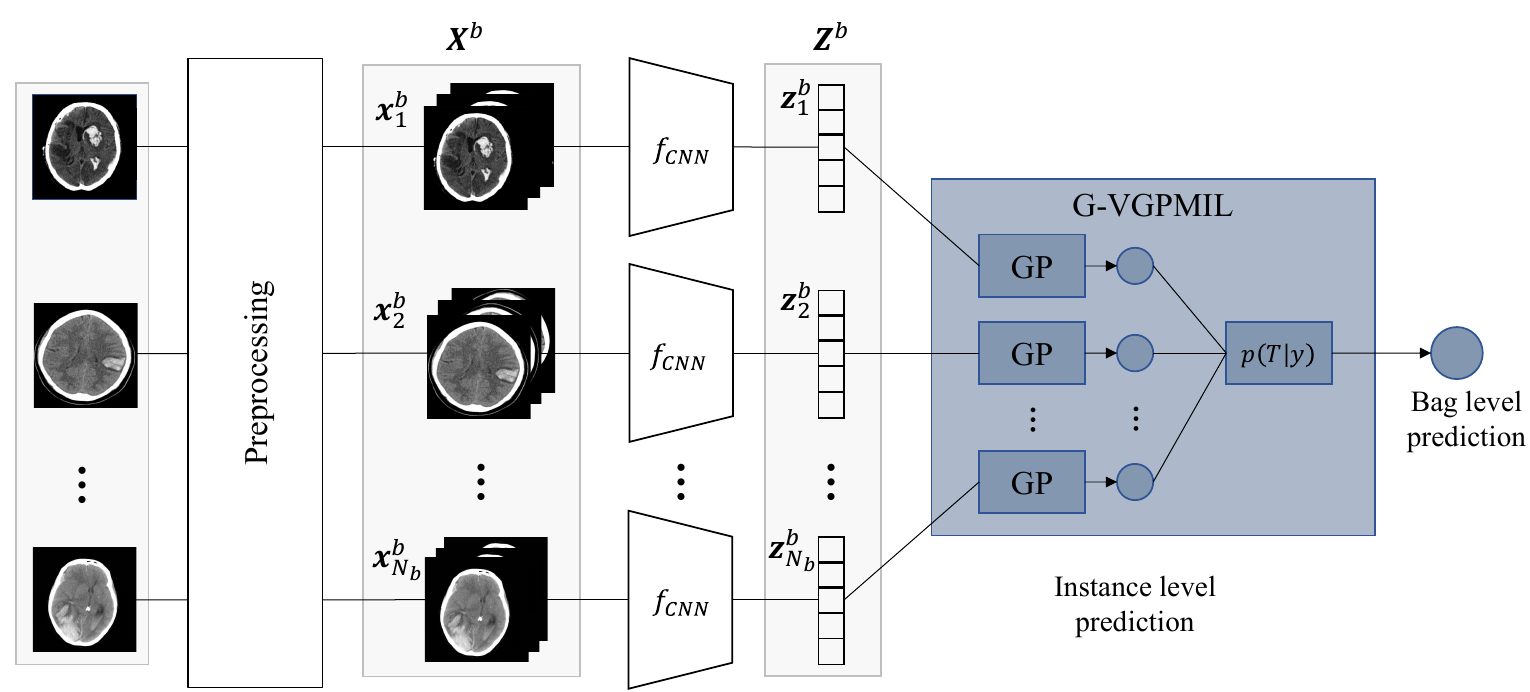}
        }
	\caption{ Attention based CNN (adapted from \cite{wu_combining_2021}).}
	\label{fig:attcnnvgpmil_model_diagram}
\end{figure}

\textbf{Preprocessing.} 
We follow the same approach as in \cite{wu_combining_2021}. 
To imitate the way radiologists read CT images we apply three windows to each CT slice to enhance the display of the brain, blood, and soft tissue. 
These three windows are concatenated to form a three-dimensional matrix and normalized to $[0,1]$ (see Figure \ref{fig:attcnnvgpmil_model_diagram}). 

\textbf{Feature extraction for the ICH problem.} 
In Sections \ref{section:gps4mil}, \ref{section:pg4gpmil} and \ref{section:gamma4gpmil} we have discussed different GP-based MIL methods, including our novel G-VGPMIL. 
Although these methods can be directly applied to tabular data, in the case of medical images (and more in general, high dimensional or highly structured data), it is convenient to extract a few meaningful features as a previous step. 
For the ICH detection problem we use an attention-based CNN that has been employed in previous work \cite{wu_combining_2021, lopez2022deep}.
The interested reader may consult \cite{wu_combining_2021} for details. 
The architecture is named AttCNN and corresponds to the composition of three functions, 
\begin{equation}
    \operatorname{AttCNN}(\bX^b) = \left( f_c \circ f_{\mathrm{Att}} \circ f_{\mathrm{CNN}}  \right) \left( \bX^b \right),
\end{equation}
where $\bX^b \in \R^{N_b \times 3 HW}$ is a matrix where the preprocessed instances of bag $b$ are collected. First, a CNN $f_{\mathrm{CNN}} \colon \R^{3HW} \to \R^{D}$ is applied to each instance in the bag
to obtain the latent $D$ dimensional representation of each instance. 
In this work, we have considered the values 8, 32, and 128 for $D$ (the number of features extracted). 
Then, an attention layer $f_{\mathrm{Att}} \colon \R^{N_b \times D} \to \R^{D}$ computes an aggregated representation of the bag. 
Lastly, a fully connected layer with a sigmoid output, $f_c \colon \R^D \to \left[0,1\right]$, is applied to obtain bag level predictions of $\p(T_b = 1 \mid \bX^b)$. 
AttCNN is trained to minimize the cross-entropy between its predictions and the true bag labels. 
Following \cite{wu_combining_2021, lopez2022deep}, after the training procedure has been completed, we remove $f_{Att}$ and $f_{c}$ and replace them with our G-VGPMIL model, obtaining the architecture represented in Figure \ref{fig:attcnnvgpmil_model_diagram}. 

\textbf{Experimental details.} We fix $H=100$ and choose the RBF kernel. \textcolor{review}{The initial kernel hyperparameters are $(v,l)=(0.5, D)$, where $D \in \left\{ 8, 32, 128 \right\}$ is the number of features extracted.} We perform a grid search considering $\left\{50, 100, 200\right\}$ for the number of inducing points, \textcolor{review}{$\left\{ 0.5, 1.0 \right\}$ for the $\alpha$ hyperparameter and $\left\{ 1.0, 2.5, 4.0 \right\}$ for the $\beta$ hyperparameter}. We use the same five train-test splits that were used in \cite{wu_combining_2021} and \cite{lopez2022deep}. To guide the training process we monitor the AUC metric in a validation subset and halt the training when it has not improved for \textcolor{review}{ten} epochs (an epoch is a complete update of the variational parameters). Based on the Bag Accuracy score, we select the best model and report the corresponding cross-validation metrics.

\subsubsection{Training and testing on RSNA}\label{subsection:results_ich_rsna}

As we have already mentioned, we first extract relevant features using AttCNN for $D \in \left\{ 8, 32, 128 \right\}$. With these features, we train the different configurations of VGPMIL and G-VGPMIL. In this section, we report and analyze the results when testing in the RSNA test splits. 

\begin{figure}
	\centering
	\subfloat{
		\includegraphics[trim={0cm 0cm 0cm 0cm},clip,width=0.6\textwidth]{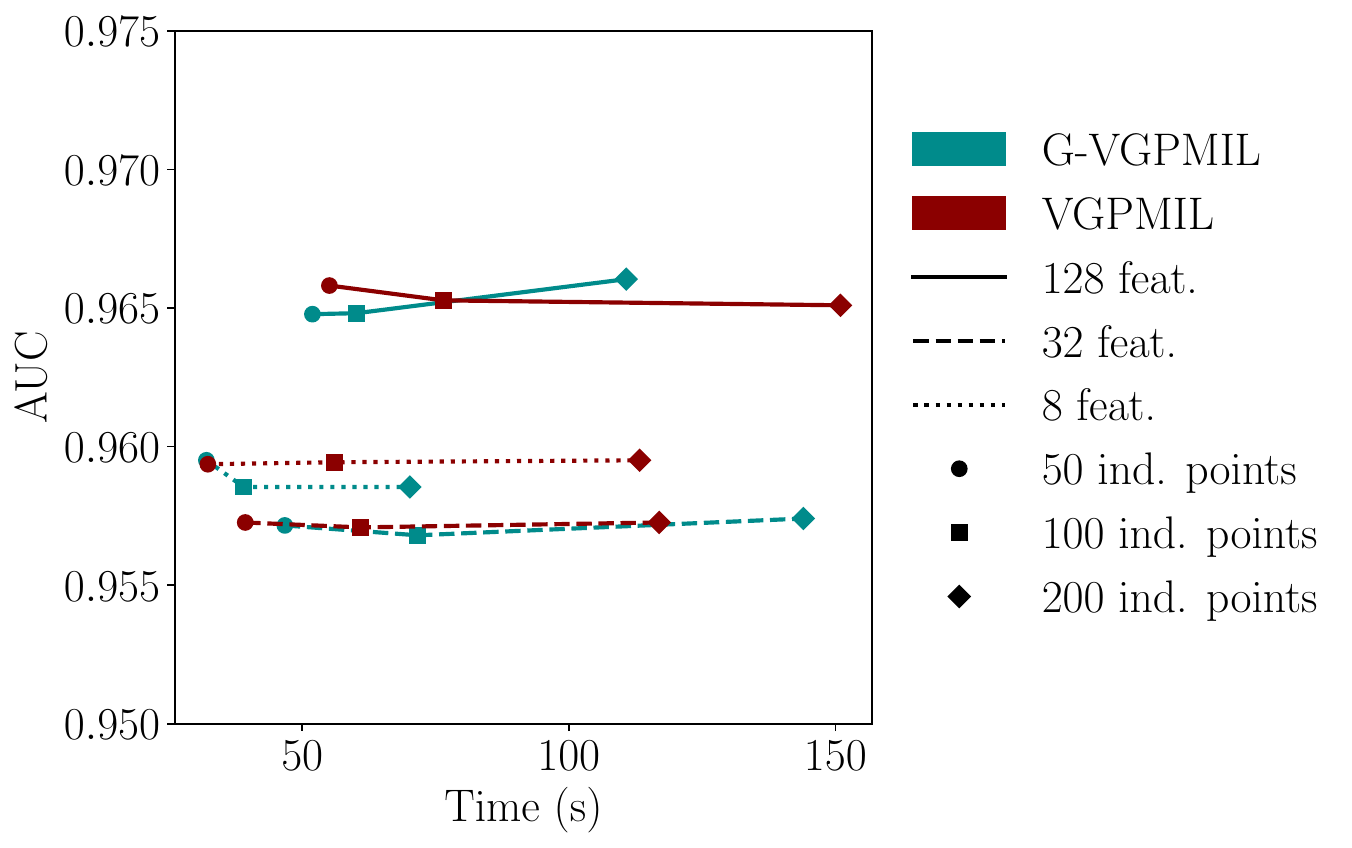}
	}
	\caption{ Training time vs AUC in the RSNA dataset. 
    }
	\label{fig:timevsauc-rsna}
\end{figure}

\textbf{Result 1: \textcolor{review}{competitive performance and training time}.} 
Figure \ref{fig:timevsauc-rsna} show an AUC vs time plot analogous to those discussed in previous experiments. 
For each value of the number of features ($D$) and inducing points, we have selected the best model according to Bag AUC. 
\textcolor{review}{In term of the bag AUC score, G-VGPMIL and VGPMIL demonstrate comparable performance, being VGPMIL slightly ahead when $D=8$. As for the training time, significant differences are present when $D \in \left\{ 32, 128 \right\}$, with G-VGPMIL showing lower training times retaining similar discriminative performance.} 

\begin{figure}[htbp]
	\centering
	\subfloat{
        \includegraphics[trim={0cm 0cm 0cm 0cm},clip,width=0.5\textwidth]{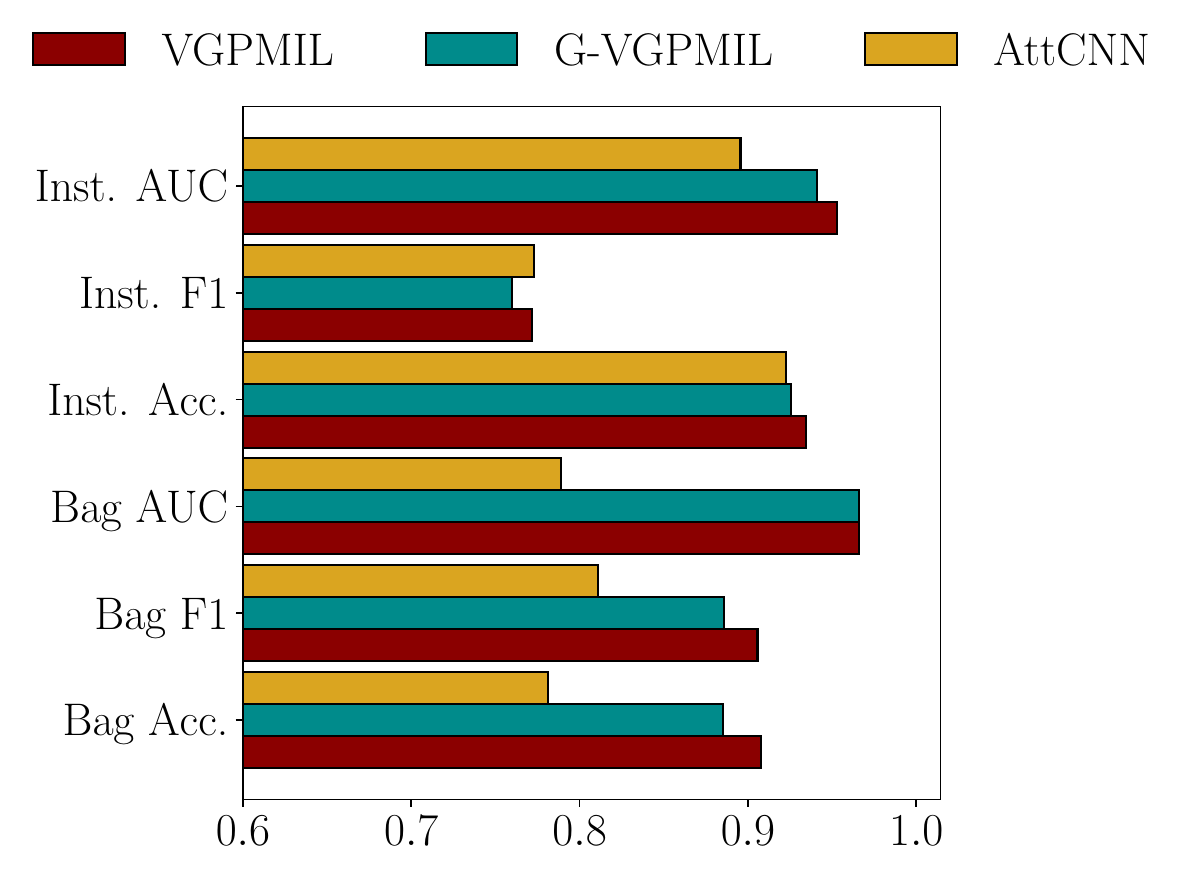
	}
        }
	\caption{ Results obtained by the best model of each type in the RSNA dataset.}
	\label{fig:bargraph-rsna}
\end{figure}

\textbf{Result 2: robustness to other metrics.} 
Once we have examined the behavior of our method using the AUC metric, we now focus on other classification metrics. 
As before, the metrics corresponding to the best model of each type (according to the Bag AUC metric) are collected in Figure \ref{fig:bargraph-rsna}. 
Notice that we are also considering AttCNN as a baseline. 
As expected, AttCNN performs much worst than the probabilistic solutions based on GPs. 
\textcolor{review}{Clearly, VGPMIL is the most effective method, being closely followed by the proposed G-VGPMIL.}
This is also true for each of the values of $D$ we have considered (see Table \ref{table:rsna}).

\begin{figure}[htbp]
	\centering
	\subfloat{
		\fbox{
        \includegraphics[trim={0cm 0cm 0cm 0cm},clip,width=0.9\textwidth]{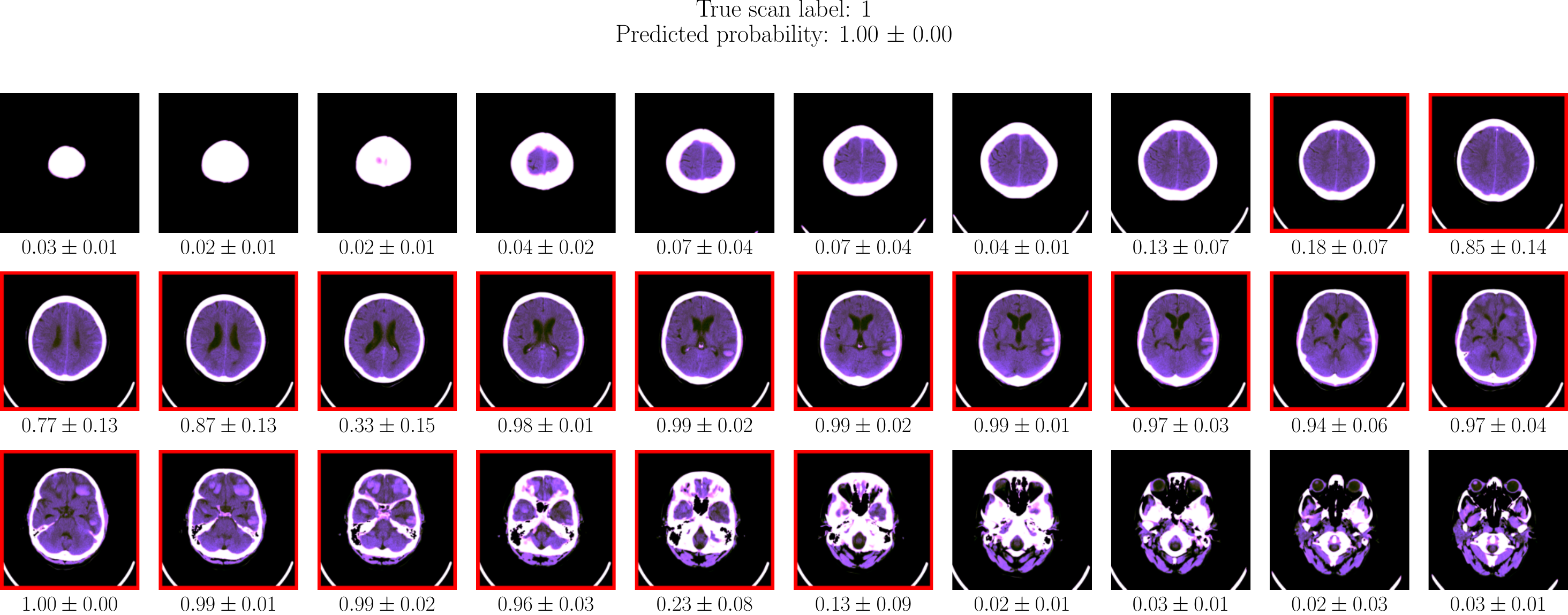}
        }
	}
	\caption{RSNA scan and G-VGPMIL predictions (probability to be positive). Positive slices are highlighted with a red frame.}
	\label{fig:rsna_real_scan_pred}
\end{figure}

\textbf{Result 3: an example of a scan prediction.} \textcolor{review}{Figure \ref{fig:rsna_real_scan_pred} shows an example of how our method predicts a positive scan from RSNA. As in Figure \ref{fig:mnist_bag_prediction_example}, the standard deviation is calculated via sampling. Our method assigns a high probability to almost all positive instances. When the presence of hemorrhage is clear, uncertainty levels (standard deviation) are very low. In slices where there is hardly any visible hemorrhage, the uncertainty is higher. Also, as a considerable number of slices are predicted as positive, uncertainty at the bag level is almost zero.} Note that, in addition to the prediction, the associated uncertainty is a very important piece of information for the user. This exemplifies the type of information that our model can provide to radiologists.


\textbf{Result 4: comparison with the state of the art.} We complete the analysis of the RSNA dataset by comparing our method to other approaches in the literature. In Table \ref{table:rsna_literature_comparison} we collect information about other methods in the literature: the size of the dataset they use, a brief description of the method, and the Bag AUC obtained when evaluating in their test set. In our case, we report the metrics obtained when evaluating RSNA. Our method outperforms the others although it uses very few scans (only \cite{sato_primitive_2018} uses less) and a relatively simple architecture (compared to 3D CNNs and Autoencoders). Also, observe that GP-based models (VGPMIL, DGPMIL, G-VGPMIL) obtain the best performance. 

\begin{table}[htbp]
	\centering
	\adjustbox{width=\textwidth}{%
		\begin{tabular}{@{}ccccc@{}}
			\toprule
			Source & Dataset size & Labeling type & Method & Bag AUC \\ \midrule
			Saab et al. \cite{shen_doubly_2019} & 4340 scans & Scan & MIL & 0.91 \\
			Jnawali et al. \cite{jnawali_deep_2018} & 40357 scans & Scan & 3D CNNs & 0.87 \\
			Titano et al. \cite{titano_automated_2018} & 37236 scans & Scan & 3D CNNs & 0.88 \\
			Sato et al. \cite{sato_primitive_2018} & 126 scans & Scan & 3D Autoencoder & 0.87 \\
			Arbabshirani et al. \cite{arbabshirani_advanced_2018} & 45583 scans & Scan & 3D CNNs & 0.85 \\
			VGPMIL (Wu et al. \cite{wu_combining_2021}) & 1150 scans & Scan & MIL & $0.9644 \pm 0.0086$ \\
            DGPMIL2 (López-Pérez et al. \cite{lopez2022deep}) & 1150 scans & Scan & MIL & $0.957$ \\
			G-VGPMIL & 1150 scans & Scan & MIL & $0.966 \pm 0.0065$ \\ \bottomrule
		\end{tabular}%
	}
	\caption{ Comparison of different approaches for binary ICH detection. VGPMIL, DGPMIL and G-VGPMIL results are obtained using the RSNA dataset for training and testing.}
    \label{table:rsna_literature_comparison}
\end{table}

\subsubsection{Evaluation on CQ500}\label{subsection:results_ich_cq500}

To finish, we evaluate our recently trained models using an external database called CQ500. 
We discuss how well our model generalizes to examples never seen before and of a different nature from the ones it was trained with.

\begin{figure}[htbp]
	\centering
	\subfloat{
		\includegraphics[trim={0cm 0cm 0cm 0cm},clip,width=0.6\textwidth]{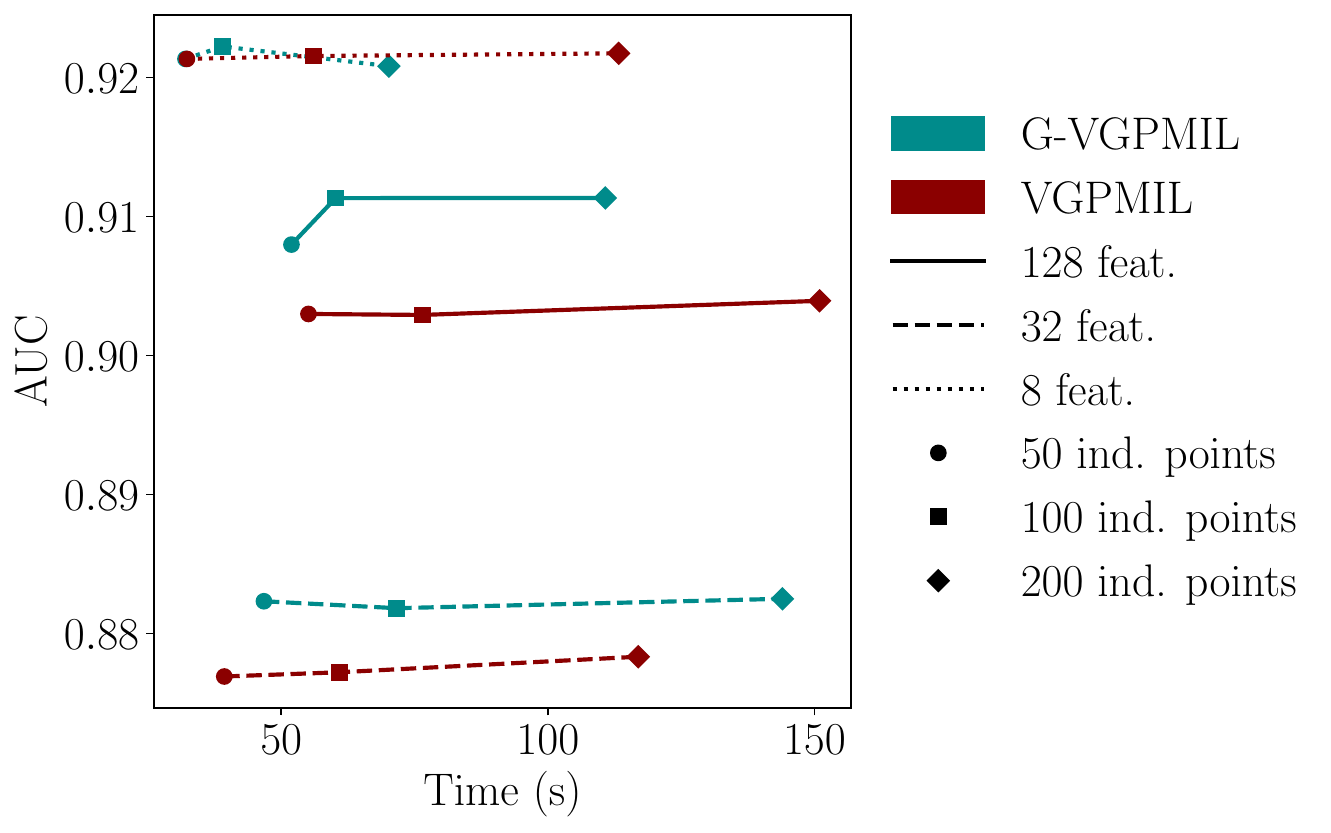}
	}
	\caption{Training time vs AUC in the CQ500 dataset.}
	\label{fig:timevsauc-cq500}
\end{figure}

\textbf{Result 1: better performance with less training time.} 
Figure \ref{fig:timevsauc-cq500} shows a plot similar to that of RSNA, now focusing on the Bag AUC in the CQ500 dataset. 
\textcolor{review}{G-VGPMIL is always above VGPMIL, which indicates better performance. Also, for $D \in \left\{ 8, 128 \right\}$, G-VGPMIL appears to the left, which implies reduced training time.}

\begin{figure}[htbp]
	\centering
	\subfloat{
        \includegraphics[trim={0cm 0cm 0cm 0cm},clip,width=0.6\textwidth]{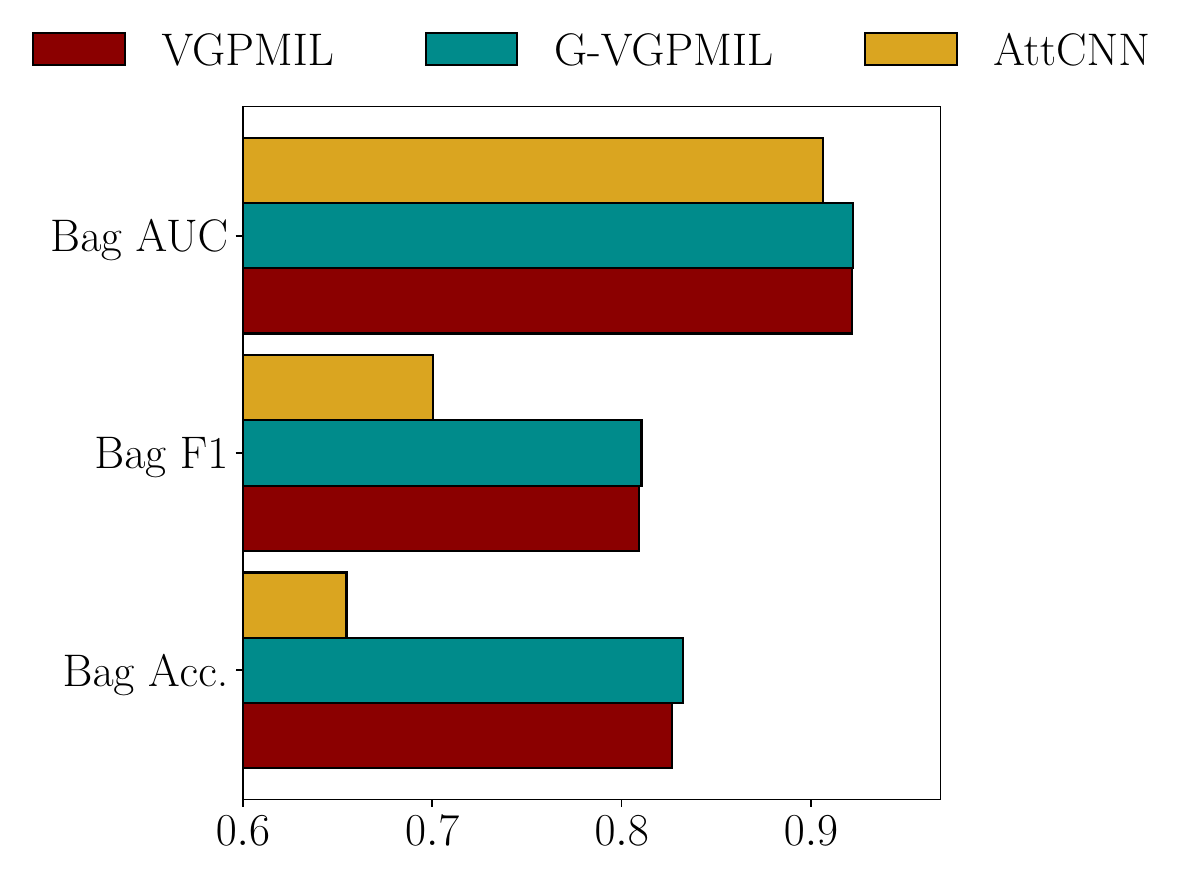
	}
        }
	\caption{ Results obtained by the best model of each type in the CQ500 dataset.}
	\label{fig:bargraph-cq500}
\end{figure}

\textbf{Result 2: other classifications metrics.} As in the case of RSNA, we report other classification metrics at bag level in Table \ref{table:cq500}. For each value of $D$, G-VGPMIL obtains the best value. In general, the gap between G-VGPMIL and VGPMIL increases as the number of used features grows. Figure \ref{fig:bargraph-cq500} represents the metrics of the best model of each type (according to Bag AUC). G-VGPMIL and VGPMIL remain at a competitive performance (being G-VGPMIL one step ahead), while the top performance of AttCNN drops significantly compared to the behavior observed in RSNA. This suggests that the generalization ability of AttCNN is worst than that of the models built upon GPs.

\textbf{Result 3: comparison with the state of the art.} Finally, we compare our approach with other methods using the Bag AUC metric in the CQ500 dataset, see Table \ref{table:cq500_literature_comparison}. Among the methods that use only bag labels to train the models \cite{monteiro_multiclass_2020, wu_combining_2021, lopez2022deep}, our method obtains the highest score. Also, it is highly competitive with those approaches that are trained using the slice labels \cite{chilamkurthy_development_2018, nguyen_cnn-lstm_2020}. Notice that using instance-level information turns this problem into a much easier one, but it requires additional effort from the radiologists. Our model is able to compete with those methods in a less demanding manner. 

\begin{table}[htbp]
\small
	\centering
	\adjustbox{width=\textwidth}{%
		\begin{tabular}{@{}cccc@{}}
			\toprule
			\multicolumn{1}{c}{Source} & \multicolumn{1}{c}{Labeling type} & \multicolumn{1}{c}{Method} & \multicolumn{1}{c}{AUC} \\ \midrule
			Chilamkurthy et al. \cite{chilamkurthy_development_2018} & Slice & 2D CNNs & 0.94 \\
			Nguyen et al. \cite{nguyen_cnn-lstm_2020} & Slice & 2D CNN + LSTM & 0.96 \\
                \midrule
			Monteiro et al. \cite{monteiro_multiclass_2020} & Scan & voxel-based CNN & 0.83 \\
			VGPMIL (Wu et al. \cite{wu_combining_2021}) & Scan & MIL & $0.9062 \pm 0.007$  \\
            DGPMIL2 (López-Pérez et al. \cite{lopez2022deep}) & Scan & MIL & $0.909$ \\
			G-VGPMIL & Scan & MIL & $0.923 \pm 0.0066$ \\ \bottomrule
		\end{tabular}%
	}
	\caption{Performance of different approaches in the CQ500 dataset. VGPMIL, DGPMIL2 and G-VGPMIL use the RSNA dataset for training.}
     \label{table:cq500_literature_comparison}
\end{table}

\section{Conclusion}\label{section:conclusion}

Motivated by the problems associated with the logistic observation model in the GP-based MIL formulation, we have reformulated VGPMIL using Pólya-Gamma random variables. We \textcolor{review}{found} that this new model, PG-VGPMIL, leads to the same update equations as VGPMIL. This is a consequence of the two equivalent representations that \textcolor{review}{the Hyperbolic Secant density} admits. This reveals that VGPMIL/PG-VGPMIL is a realization of a more general framework, \textcolor{review}{$\psi$-VGPMIL, which is achieved by replacing the Hyperbolic Secant density by a general GSM density $\psi$. }

An interesting challenge that arises is the choice of a convenient \textcolor{review}{density}. In this work, we have explored the natural choice of the Gamma distribution, arriving at the newly proposed G-VGPMIL. Our experiments show that G-VGPMIL improves upon VGPMIL in terms of predictive performance and training time. G-VGPMIL shows competitive results with fully supervised models, thus closing the gap with them. As the features used to train G-VGPMIL may not be optimal (they have been extracted in two phases), we believe that there is still room for improvement. 

{\color{review}
Another interesting challenge is the extension of our model to a multi-class problem, where one has to deal with the intractable term that the softmax function introduces. Following the line of our work, we would need to express the partition function of the softmax as a GSM, which is not obvious and requires further investigation. 
}

Finally, we believe that this work provides interesting ideas that can be useful beyond MIL and leaves open questions of both theoretical and practical nature. We hope that our ideas will be useful to the rest of the community and can enhance further research. 


\section*{Acknowledgments}
This work was supported by project PID2019-105142RB-C22 funded by MCIN, by project PID2022-140189OB-C22 funded by MCIN / AEI / 10.13039 / 501100011033, and by project B‐TIC‐324‐UGR20 funded by FEDER/Junta de Andalucía and Universidad de Granada. Pablo Morales-Álvarez acknowledges grant C-EXP-153-UGR23 funded by Consejería de Universidad,
Investigación e Innovación and by ERDF Andalusia Program 2021-2027. The work by Francisco M. Castro-Macías was supported by Ministerio de Universidades under FPU contract FPU21/01874. The authors acknowledge University of Granada / CBUA for funding the open access charge. 


\bibliographystyle{elsarticle-num} 
\bibliography{ref}

\newpage

\begin{appendices}

\section{$\psi$-VGPMIL variational updates}\label{appendix:variational_updates}

In this section, we derive the variational updates for the $\psi$-VGPMIL model presented in Section \ref{section:general_model}. The updates for PG-VGPMIL, from Section \ref{section:pg4gpmil}, are obtained by setting $\psi$ to the Hyperbolic Secant distribution defined by Eq. \eqref{eq:obs_model_phi}. Similarly, for G-VGPMIL, as discussed in Section \ref{section:gamma4gpmil}, the updates are obtained by defining $\psi$ as the density specified in Eq. \eqref{eq:psi_gamma}. 

To perform inference, we consider the augmented $\psi$-VGPMIL model. Recall that $\psi$ is a GSM density, so there exists a mixing density $\widehat{\psi} \colon \left[ 0, +\infty \right[ \to \R$ such that, 
\begin{equation}\label{eq:psi_gsm}
    \psi(x) = \int_{0}^{+\infty} \mathcal{N}\left( x \mid 0, \omega^{-1} \right) \widehat{\psi}(\omega) \dd \omega, \quad \forall x \in \R.
\end{equation}
The augmented $\psi$-VGPMIL model is given by
\begin{gather}
    \p(\bu, \bff, \by, \bomega, \bT) = \p(\bu, \bff, \by, \bomega) \p(\bT \mid \by), \\
    \p(\bu, \bff, \by, \bomega) = Z^{-1} \mathcal{N}\left( \bff, \bu \right) \exp\left(  (\by - 2^{-1}\mathbf{1})^\top \bff \right) \mathcal{N}\left( \bff \mid \bomega \right) \widehat{\psi}(\bomega),
\end{gather}
where $Z = \pi^{-N} \int \mathcal{N}\left( \bff, \bu \right) \psi(\bff) \phi(\bff)^{-1} \dd (\bff, \bu)$, $\widehat{\psi}(\bomega) = \prod_{n=1}^N \widehat{\psi}(\omega_n)$, $\mathcal{N}\left( \bff \mid \bomega \right) = \mathcal{N}\left( \bff \mid \bzero, \bOmega^{-1} \right)$, $\bOmega = \operatorname{Diag}\left( \bomega \right)$, and $\p(\bT \mid \by)$ is the bag likelihood given by Eq. \eqref{eq:bag_prob}. 

Following the variational inference approach, we approximate the posterior $\p(\bu, \bff, \by, \bomega \mid \bT)$ with a variational distribution $\q(\bu, \bff, \by, \bomega) = \q(\bu) \mathcal{N}(\bff \mid \bu) \q(\by) \q(\bomega)$, where $\q(\by) = \prod_{n=1}^N\q(y_n)$ and $\q(\bomega) = \prod_{n=1}^N\q(\omega_n)$. We achieve this by minimizing the KL divergence between the posterior and the variational distributions. Using \cite[Eq. (10.9)]{bishop2006pattern}, we can compute the optimal solution for each distribution keeping the others fixed,
\begin{equation}
	\log \q = \Ebb_{\Q \setminus \q}\left[ \log \p(\bT, \bu, \bff, \by, \bomega) \right] + \const,
\end{equation}
where $\Q$ denotes the joint variational distribution.

\textbf{Update of $\q(y_n)$.} We denote $\q(y_{j \neq n}) = \prod_{j \neq n} \pi_j^{y_j} (1-\pi_j)^{y_j}$. Let $n$ be fixed and let $b$ be the index of the bag to which instance $n$ belongs. We have
\begin{align}
    \log \q(y_n) = & \underbrace{\Ebb_{\q(\bu) \mathcal{N}(\bff \mid \bu) \q(y_{j \neq n})} \left[ \left( \by - 2^{-1}\mathbf{1} \right)^{\top} \bff \right]}_{A_1} + \\
    & + \underbrace{\Ebb_{\q(y_{j \neq n})} \left[ \log \p(\bT \mid \by) \right]}_{A_2} + \const.
\end{align}
We analyze each term separately, 
\begin{align}
	A_1 & = \Ebb_{\q(y_{j \neq n})} \left[ \by \right]^\top \Ebb_{\q(\bu) \p(\bff \mid \bu)} \left[ \bff \right] + \const \\
	& = y_n\Ebb_{\q(\bu) \p(\bff \mid \bu)} \left[ f_n \right] + \const = \\
	& = y_n\bK_{\bx_i \bZ} \bK_{\bZ \bZ}^{-1} \bm + \const \\
	A_2 & =  \log H  \Ebb_{\q(y_{j \neq n})} \left[ G_b \right] + \const =  \\
	& = \log H (2 T_b - 1) \Ebb_{\q(y_{j \neq n})} \left[ \max \left\{ \by \right\}_b \right] + \const = \\
	& = y_n \log H (2 T_b - 1)  \left( 1 - \Ebb \left[ \max \left\{ \by \right\}_{ b\setminus n} \right] \right) + \const,
\end{align}
where we have used \cite[Eq. 13]{haussmann2017variational} to compute the expected value involved. Putting it all together, 
\begin{align}
	\log \q(y_n) = y_n \left[ \bK_{\bx_i \bZ} \bK_{\bZ \bZ}^{-1} \bm + \log H (2 T_b - 1)  \left( 1 - \Ebb \left[ \max \left\{ \by \right\}_{ b\setminus n} \right] \right) \right] + \const,
\end{align}
from which we deduce $\q(y_n) = \operatorname{Bernouilli}\left( y_n \mid \pi_n \right)$ with
\begin{align}
	\pi_n = \logit \left[ \bK_{\bx_i \bZ} \bK_{\bZ \bZ}^{-1} \bm + \log H (2 T_b - 1)  \left( 1 - \Ebb \left[ \max \left\{ \by \right\}_{ b\setminus n} \right] \right) \right]
\end{align}	
\textbf{Update of $\q(\bu)$.}
\begin{equation}
	\log \q(\bu) = \underbrace{\log \mathcal{N}(\bu)}_{B_1} + \underbrace{ \Ebb_{ \mathcal{N}(\bff \mid \bu) \q(\by) \q(\bomega)} \left[ \left(  (\by - 2^{-1}\mathbf{1})^\top \bff \right) + \log \mathcal{N}\left( \bff \mid \bomega \right) \right] }_{B_2} + \const.
\end{equation}
We analyze each term separately,
\begin{align}
	B_1 = & - 2^{-1} \bu^\top \bK_{\bZ \bZ}^{-1} \bu + \const, \\
	B_2 = & \Ebb_{\mathcal{N}(\bff \mid \bu) \q(\by) \q(\bomega)} \left[ \left(\by - 1/2\right)^{\top}\bff - 2^{-1}\bff^{\top} \bOmega \bff \right] + \const = \\
	= & \Ebb_{\mathcal{N}(\bff \mid \bu) \q(\bomega)} \left[ \left( \bpi - 2^{-1}\mathbf{1} \right)^{\top}\bff - 2^{-1}\bff^{\top} \bOmega \bff \right] + \const = \\
	= & \left( \bpi - 2^{-1}\mathbf{1} \right)^{\top} \bK_{\bX \bZ} \bK_{\bZ \bZ}^{-1}\bu - 2^{-1} \bu^{\top} \bK_{\bZ \bZ}^{-1} \bK_{\bZ \bX} \bTheta \bK_{\bX \bZ} \bK_{\bZ \bZ}^{-1} \bu  + \const,
\end{align}
where $\bTheta = \diag(\theta_1, \ldots, \theta_N)$ and $\theta_n = \Ebb_{\q(\omega_n)}\left[ \omega_n \right]$. Putting all together
\begin{align}
	\log \q(\bu) = & \left( \bpi - 2^{-1}\mathbf{1} \right)^{\top} \bK_{\bX \bZ} \bK_{\bZ \bZ}^{-1}\bu \ + \\
    & - 2^{-1} \bu^\top \left( \bK_{\bZ \bZ}^{-1} \bK_{\bZ \bX} \bTheta \bK_{\bX \bZ} \bK_{\bZ \bZ}^{-1} + \bK_{\bZ \bZ}^{-1} \right) \bu + \const.
\end{align}
This proves that $\q(\bu) = \mathcal{N}(\bu \mid \bm, \bS)$ with
\begin{gather}
	\bm = \bS\bK_{\bZ \bZ}^{-1} \bK_{\bZ \bX}  \left( \bpi - 2^{-1}\mathbf{1} \right)\\
	\bS = \left( \bK_{\bZ \bZ}^{-1} \bK_{\bZ \bX} \bTheta  \bK_{\bX \bZ} \bK_{\bZ \bZ}^{-1} + \bK_{\bZ \bZ}^{-1} \right)^{-1}
\end{gather}
\textbf{Update of $\q(\omega_n)$.} Let $n$ be fixed. 
\begin{align}
    \log \q(\omega_n) & = \Ebb_{\q(f_n)}\left[ \log \mathcal{N}\left( f_n \mid 0, \omega_n^{-1}\right) \right] + \log \widehat{\psi}(\omega_n) = \\
    & = \log \mathcal{N}\left( c_n \mid 0, \omega_n^{-1}\right) + \log \widehat{\psi}(\omega_n),
\end{align}
where $c_n = \sqrt{ \Ebb_{\q(f_n)}\left[ f_n^2 \right] }$. Thus, if $\psi(c_n) \neq 0$, 
\begin{equation}\label{eq:app_q_omega}
    \q(\omega_n) = \psi(c_n)^{-1} \mathcal{N}\left( c_n \mid 0, \omega_n^{-1}\right) \widehat{\psi}(\omega_n).
\end{equation}
In the PG-VGPMIL scenario, we can deduce $\q(\omega_n) = \operatorname{PG}\left( \omega_n \mid 1, c_n \right)$ using that the general Pólya-Gamma density $\operatorname{PG}\left( \cdot \mid b, c\right)$ is a exponential tilting of the $\operatorname{PG}\left( \cdot \mid b, 0\right)$ density \cite{polson2013bayesian}. However, in the general case, we only need the expectection of $\q(\omega_n)$. To calculate it, we first differentiate under the integral sign in Eq. \eqref{eq:psi_gsm}, 
\begin{equation}
    \psi'(x) = -x \int_{0}^{+\infty} \omega \mathcal{N}\left( x \mid 0, \omega^{-1} \right) \widehat{\psi}(\omega) \dd \omega.
\end{equation}
Then, we evaluate this expression in $x=c_n$ and use Eq. \eqref{eq:app_q_omega}, 
\begin{equation}
    \psi'(c_n) = - c_n \psi(c_n) \Ebb_{\q(\omega_n)}\left[ \omega_n\right],
\end{equation}
from where we conclude that $\theta_n = -\psi'(c_n)/(c_n \psi(c_n))$.

\section{Tables}\label{appendix:tables}

The top value of each metric within each group is highlighted in bold, and the best in each column is underlined.

\begin{table}[H]
	\centering
	\begin{adjustbox}{width=\textwidth}
		\begin{tabular}{cccccccc}
			\toprule
			Num. ind. points & Model & Inst. Accuracy & Inst. F1 & Inst. AUC & Bag Accuracy & Bag F1 & Bag AUC \\
			\midrule
			\multirow[c]{2}{*}{50} & G-VGPMIL & $\mathbf{0.9342 \pm 0.004}$ & $\mathbf{0.6769 \pm 0.0305}$ & $0.9232 \pm 0.0056$ & $\mathbf{0.8481 \pm 0.0153}$ & $\mathbf{0.843 \pm 0.018}$ & $\mathbf{0.9166 \pm 0.0136}$ \\
			& VGPMIL & $0.9296 \pm 0.0013$ & $0.6352 \pm 0.0083$ & $\mathbf{0.9246 \pm 0.0071}$ & $0.8438 \pm 0.011$ & $0.8327 \pm 0.0117$ & $0.9125 \pm 0.0098$ \\
			\cline{1-8}
			\multirow[c]{2}{*}{100} & G-VGPMIL & $\mathbf{0.9431 \pm 0.0029}$ & $\mathbf{0.7183 \pm 0.0178}$ & $\mathbf{0.9559 \pm 0.0032}$ & $\mathbf{0.8841 \pm 0.0139}$ & $\mathbf{0.878 \pm 0.0151}$ & $\mathbf{0.947 \pm 0.0069}$ \\
			& VGPMIL & $0.9425 \pm 0.0033$ & $0.7171 \pm 0.0209$ & $0.95 \pm 0.0032$ & $0.8756 \pm 0.0188$ & $0.8699 \pm 0.0203$ & $0.9414 \pm 0.0071$ \\
			\cline{1-8}
			\multirow[c]{2}{*}{200} & G-VGPMIL & $0.9563 \pm 0.0023$ & $0.7962 \pm 0.0137$ & $\underline{\mathbf{0.972 \pm 0.0009}}$ & $0.9131 \pm 0.0132$ & $0.911 \pm 0.014$ & $\underline{\mathbf{0.9656 \pm 0.003}}$ \\
			& VGPMIL & $\underline{\mathbf{0.957 \pm 0.004}}$ & $\underline{\mathbf{0.8006 \pm 0.0247}}$ & $0.9695 \pm 0.002$ & $\underline{\mathbf{0.9147 \pm 0.0137}}$ & $\underline{\mathbf{0.9133 \pm 0.0148}}$ & $0.9654 \pm 0.0056$ \\
			\bottomrule
		\end{tabular}
	\end{adjustbox}
	\caption{\centering MNIST PCA }
	\label{table:mnist_pca}
\end{table}

\begin{table}[H]
	\centering
	\begin{adjustbox}{width=\textwidth}
		\begin{tabular}{cccccccc}
			\toprule
			Num. ind. points & Model & Inst. Accuracy & Inst. F1 & Inst. AUC & Bag Accuracy & Bag F1 & Bag AUC \\
			\midrule
			\multirow[c]{2}{*}{50} & G-VGPMIL & $\mathbf{0.9016 \pm 0.0036}$ & $\mathbf{0.5365 \pm 0.0448}$ & $\mathbf{0.8703 \pm 0.015}$ & $\mathbf{0.7384 \pm 0.0193}$ & $\mathbf{0.75 \pm 0.0274}$ & $\mathbf{0.8355 \pm 0.0234}$ \\
			& VGPMIL & $0.8961 \pm 0.0039$ & $0.3636 \pm 0.0346$ & $0.8549 \pm 0.0094$ & $0.7016 \pm 0.0189$ & $0.6311 \pm 0.0301$ & $0.8035 \pm 0.0168$ \\
			\cline{1-8}
			\multirow[c]{2}{*}{100} & G-VGPMIL & $\mathbf{0.9113 \pm 0.0018}$ & $\mathbf{0.5681 \pm 0.0147}$ & $\mathbf{0.8827 \pm 0.0056}$ & $\mathbf{0.7653 \pm 0.0147}$ & $\mathbf{0.771 \pm 0.0141}$ & $\mathbf{0.8581 \pm 0.0114}$ \\
			& VGPMIL & $0.9052 \pm 0.0016$ & $0.4397 \pm 0.0224$ & $0.8737 \pm 0.0059$ & $0.7522 \pm 0.0074$ & $0.705 \pm 0.0154$ & $0.8408 \pm 0.0079$ \\
			\cline{1-8}
			\multirow[c]{2}{*}{200} & G-VGPMIL & $\underline{\mathbf{0.916 \pm 0.0024}}$ & $\underline{\mathbf{0.5764 \pm 0.0241}}$ & $\underline{\mathbf{0.8835 \pm 0.0096}}$ & $\underline{\mathbf{0.7822 \pm 0.0114}}$ & $\underline{\mathbf{0.7792 \pm 0.016}}$ & $\underline{\mathbf{0.863 \pm 0.0152}}$ \\
			& VGPMIL & $0.9051 \pm 0.0016$ & $0.4338 \pm 0.0107$ & $0.876 \pm 0.0065$ & $0.7491 \pm 0.0075$ & $0.6986 \pm 0.0099$ & $0.8476 \pm 0.0082$ \\
			\bottomrule
		\end{tabular}
	\end{adjustbox}
	\caption{\centering MNIST RAW}
	\label{table:mnist_raw}
\end{table}

\begin{table}[H]
	\centering
	\begin{adjustbox}{width=\textwidth}
		\begin{tabular}{ccccc}
			\toprule
			Num. ind. points & Model & Bag Accuracy & Bag F1 & Bag AUC \\
			\midrule
			\multirow[c]{2}{*}{50} & G-VGPMIL & $0.8832 \pm 0.0202$ & $\mathbf{0.8917 \pm 0.0182}$ & $\mathbf{0.9607 \pm 0.0133}$ \\
			& VGPMIL & $\mathbf{0.8832 \pm 0.018}$ & $0.8916 \pm 0.0175$ & $0.958 \pm 0.0136$ \\
			\cline{1-5}
			\multirow[c]{2}{*}{100} & G-VGPMIL & $\underline{\mathbf{0.905 \pm 0.0224}}$ & $\mathbf{0.9078 \pm 0.0209}$ & $\underline{\mathbf{0.9711 \pm 0.016}}$ \\
			& VGPMIL & $0.8886 \pm 0.0178$ & $0.8956 \pm 0.0171$ & $0.9682 \pm 0.0137$ \\
			\cline{1-5}
			\multirow[c]{2}{*}{200} & G-VGPMIL & $\mathbf{0.9023 \pm 0.0258}$ & $\underline{\mathbf{0.908 \pm 0.0233}}$ & $\mathbf{0.9669 \pm 0.0159}$ \\
			& VGPMIL & $0.8806 \pm 0.0271$ & $0.89 \pm 0.0218$ & $0.9617 \pm 0.0141$ \\
			\bottomrule
		\end{tabular}
	\end{adjustbox}
	\caption{\centering MUSK1 }
	\label{table:musk1}
\end{table}

\begin{table}[H]
	\centering
	\begin{adjustbox}{width=\textwidth}
		\begin{tabular}{ccccc}
			\toprule
			Num. ind. points & Model & Bag Accuracy & Bag F1 & Bag AUC \\
			\midrule
			\multirow[c]{2}{*}{50} & G-VGPMIL & $0.8308 \pm 0.0298$ & $\mathbf{0.7806 \pm 0.0382}$ & $\mathbf{0.9006 \pm 0.0247}$ \\
			& VGPMIL & $\mathbf{0.8358 \pm 0.0412}$ & $0.779 \pm 0.0609$ & $0.8978 \pm 0.0219$ \\
			\cline{1-5}
			\multirow[c]{2}{*}{100} & G-VGPMIL & $\mathbf{0.8603 \pm 0.021}$ & $\mathbf{0.8228 \pm 0.0174}$ & $\mathbf{0.9281 \pm 0.0228}$ \\
			& VGPMIL & $0.853 \pm 0.0316$ & $0.797 \pm 0.0492$ & $0.923 \pm 0.0303$ \\
			\cline{1-5}
			\multirow[c]{2}{*}{200} & G-VGPMIL & $\underline{\mathbf{0.8971 \pm 0.0293}}$ & $\underline{\mathbf{0.8617 \pm 0.0385}}$ & $\underline{\mathbf{0.9605 \pm 0.01}}$ \\
			& VGPMIL & $0.88 \pm 0.0234$ & $0.834 \pm 0.0365$ & $0.9488 \pm 0.0156$ \\
			\bottomrule
		\end{tabular}
	\end{adjustbox}
	\caption{\centering MUSK2}
	\label{table:musk2}
\end{table}

\begin{table}[H]
	\centering
	\begin{adjustbox}{width=\textwidth}
		\begin{tabular}{ccccccccc}
			\toprule
			Num. features & Num. ind. points & Model & RSNA inst. Accuracy & RSNA inst. F1 & RSNA inst. AUC & RSNA bag Accuracy & RSNA bag F1 & RSNA bag AUC \\
			\midrule
			\multirow[c]{6}{*}{8} & \multirow[c]{2}{*}{50} & G-VGPMIL & $0.9395 \pm 0.002$ & $0.7791 \pm 0.0125$ & $0.9508 \pm 0.0053$ & $\mathbf{0.8907 \pm 0.0172}$ & $\mathbf{0.8855 \pm 0.0154}$ & $\mathbf{0.9595 \pm 0.0081}$ \\
			&  & VGPMIL & $\mathbf{0.941 \pm 0.0016}$ & $\underline{\mathbf{0.7926 \pm 0.0054}}$ & $\mathbf{0.9592 \pm 0.0034}$ & $0.884 \pm 0.01$ & $0.8823 \pm 0.0095$ & $0.9594 \pm 0.0075$ \\
			\cline{2-9}
			& \multirow[c]{2}{*}{100} & G-VGPMIL & $0.9399 \pm 0.0018$ & $0.7823 \pm 0.0123$ & $0.9434 \pm 0.0045$ & $0.8867 \pm 0.0202$ & $0.8818 \pm 0.019$ & $0.9585 \pm 0.008$ \\
			&  & VGPMIL & $\mathbf{0.941 \pm 0.0015}$ & $\mathbf{0.7903 \pm 0.0049}$ & $\mathbf{0.9597 \pm 0.0036}$ & $\mathbf{0.888 \pm 0.016}$ & $\mathbf{0.8851 \pm 0.0156}$ & $\mathbf{0.9594 \pm 0.0076}$ \\
			\cline{2-9}
			& \multirow[c]{2}{*}{200} & G-VGPMIL & $0.9401 \pm 0.0018$ & $0.7835 \pm 0.0109$ & $0.9408 \pm 0.0052$ & $0.8867 \pm 0.0174$ & $0.882 \pm 0.0164$ & $0.9585 \pm 0.0079$ \\
			&  & VGPMIL & $\underline{\mathbf{0.9411 \pm 0.0013}}$ & $\mathbf{0.7875 \pm 0.0041}$ & $\underline{\mathbf{0.9603 \pm 0.0037}}$ & $\mathbf{0.8933 \pm 0.0207}$ & $\mathbf{0.8883 \pm 0.0211}$ & $\mathbf{0.9595 \pm 0.0074}$ \\
			\cline{1-9} \cline{2-9}
			\multirow[c]{6}{*}{32} & \multirow[c]{2}{*}{50} & G-VGPMIL & $0.9251 \pm 0.0103$ & $0.7554 \pm 0.0307$ & $0.9418 \pm 0.0078$ & $0.8613 \pm 0.0229$ & $0.8648 \pm 0.0186$ & $0.9572 \pm 0.0044$ \\
			&  & VGPMIL & $\mathbf{0.9337 \pm 0.008}$ & $\mathbf{0.7647 \pm 0.0292}$ & $\mathbf{0.9523 \pm 0.0054}$ & $\mathbf{0.8933 \pm 0.0189}$ & $\mathbf{0.8904 \pm 0.0176}$ & $\mathbf{0.9573 \pm 0.0049}$ \\
			\cline{2-9}
			& \multirow[c]{2}{*}{100} & G-VGPMIL & $0.9254 \pm 0.0105$ & $0.7563 \pm 0.0311$ & $0.9425 \pm 0.0075$ & $0.86 \pm 0.0276$ & $0.8641 \pm 0.0239$ & $0.9568 \pm 0.0053$ \\
			&  & VGPMIL & $\mathbf{0.9329 \pm 0.008}$ & $\mathbf{0.76 \pm 0.0292}$ & $\mathbf{0.9527 \pm 0.0053}$ & $\mathbf{0.8907 \pm 0.0182}$ & $\mathbf{0.8873 \pm 0.017}$ & $\mathbf{0.9571 \pm 0.0053}$ \\
			\cline{2-9}
			& \multirow[c]{2}{*}{200} & G-VGPMIL & $0.9266 \pm 0.012$ & $0.7562 \pm 0.0344$ & $0.9426 \pm 0.0074$ & $0.864 \pm 0.0248$ & $0.8657 \pm 0.0204$ & $\mathbf{0.9574 \pm 0.005}$ \\
			&  & VGPMIL & $\mathbf{0.9325 \pm 0.0079}$ & $\mathbf{0.7568 \pm 0.0287}$ & $\mathbf{0.9531 \pm 0.0052}$ & $\mathbf{0.8893 \pm 0.0187}$ & $\mathbf{0.8854 \pm 0.018}$ & $0.9573 \pm 0.0049$ \\
			\cline{1-9} \cline{2-9}
			\multirow[c]{6}{*}{128} & \multirow[c]{2}{*}{50} & G-VGPMIL & $0.9245 \pm 0.0069$ & $0.7586 \pm 0.0162$ & $0.9407 \pm 0.0044$ & $0.8773 \pm 0.0124$ & $0.8793 \pm 0.0119$ & $0.9648 \pm 0.0073$ \\
			&  & VGPMIL & $\mathbf{0.9347 \pm 0.007}$ & $\mathbf{0.7716 \pm 0.0217}$ & $\mathbf{0.9529 \pm 0.0047}$ & $\mathbf{0.908 \pm 0.0154}$ & $\mathbf{0.9057 \pm 0.0151}$ & $\underline{\mathbf{0.9658 \pm 0.0067}}$ \\
			\cline{2-9}
			& \multirow[c]{2}{*}{100} & G-VGPMIL & $0.9276 \pm 0.0057$ & $0.7509 \pm 0.0178$ & $0.9412 \pm 0.0061$ & $0.9067 \pm 0.0112$ & $0.9029 \pm 0.0121$ & $0.9648 \pm 0.0075$ \\
			&  & VGPMIL & $\mathbf{0.9353 \pm 0.007}$ & $\mathbf{0.7721 \pm 0.0219}$ & $\mathbf{0.9532 \pm 0.0046}$ & $\mathbf{0.912 \pm 0.0129}$ & $\mathbf{0.9089 \pm 0.013}$ & $\mathbf{0.9653 \pm 0.0067}$ \\
			\cline{2-9}
			& \multirow[c]{2}{*}{200} & G-VGPMIL & $0.9258 \pm 0.0068$ & $0.7601 \pm 0.0171$ & $0.9409 \pm 0.0047$ & $0.8853 \pm 0.0181$ & $0.8858 \pm 0.0168$ & $\mathbf{0.9652 \pm 0.0065}$ \\
			&  & VGPMIL & $\mathbf{0.935 \pm 0.0079}$ & $\mathbf{0.7683 \pm 0.0254}$ & $\mathbf{0.954 \pm 0.0047}$ & $\underline{\mathbf{0.9133 \pm 0.0184}}$ & $\underline{\mathbf{0.9097 \pm 0.0186}}$ & $0.9651 \pm 0.0068$ \\
			\bottomrule
		\end{tabular}
	\end{adjustbox}
	\caption{\centering RSNA}
	\label{table:rsna}
\end{table}

\begin{table}[H]
	\centering
	\begin{adjustbox}{width=\textwidth}
		\begin{tabular}{cccccc}
			\toprule
			Num. features & Num. ind. points & Model & CQ500 bag Accuracy & CQ500 bag F1 & CQ500 bag AUC \\
			\midrule
			\multirow[c]{6}{*}{8} & \multirow[c]{2}{*}{50} & G-VGPMIL & $\mathbf{0.831 \pm 0.0115}$ & $\mathbf{0.8115 \pm 0.0108}$ & $\mathbf{0.9213 \pm 0.0058}$ \\
			&  & VGPMIL & $0.8041 \pm 0.0131$ & $0.7923 \pm 0.0122$ & $0.9213 \pm 0.0063$ \\
			\cline{2-6}
			& \multirow[c]{2}{*}{100} & G-VGPMIL & $\mathbf{0.8322 \pm 0.0065}$ & $\mathbf{0.8105 \pm 0.0073}$ & $\underline{\mathbf{0.923 \pm 0.0066}}$ \\
			&  & VGPMIL & $0.811 \pm 0.0117$ & $0.7966 \pm 0.0109$ & $0.9215 \pm 0.0062$ \\
			\cline{2-6}
			& \multirow[c]{2}{*}{200} & G-VGPMIL & $\mathbf{0.831 \pm 0.0093}$ & $\mathbf{0.8103 \pm 0.0079}$ & $0.9208 \pm 0.0058$ \\
			&  & VGPMIL & $0.8265 \pm 0.0082$ & $0.8092 \pm 0.0077$ & $\mathbf{0.9217 \pm 0.0062}$ \\
			\cline{1-6} \cline{2-6}
			\multirow[c]{6}{*}{32} & \multirow[c]{2}{*}{50} & G-VGPMIL & $\mathbf{0.8233 \pm 0.0271}$ & $\mathbf{0.79 \pm 0.0394}$ & $\mathbf{0.8823 \pm 0.0284}$ \\
			&  & VGPMIL & $0.8022 \pm 0.0273$ & $0.7752 \pm 0.0377$ & $0.8769 \pm 0.0298$ \\
			\cline{2-6}
			& \multirow[c]{2}{*}{100} & G-VGPMIL & $\mathbf{0.8245 \pm 0.0263}$ & $\mathbf{0.7904 \pm 0.0391}$ & $\mathbf{0.8818 \pm 0.028}$ \\
			&  & VGPMIL & $0.8063 \pm 0.0284$ & $0.7778 \pm 0.0384$ & $0.8772 \pm 0.0294$ \\
			\cline{2-6}
			& \multirow[c]{2}{*}{200} & G-VGPMIL & $\mathbf{0.829 \pm 0.023}$ & $\mathbf{0.7927 \pm 0.0385}$ & $\mathbf{0.8825 \pm 0.0288}$ \\
			&  & VGPMIL & $0.8127 \pm 0.0309$ & $0.7829 \pm 0.0423$ & $0.8783 \pm 0.0293$ \\
			\cline{1-6} \cline{2-6}
			\multirow[c]{6}{*}{128} & \multirow[c]{2}{*}{50} & G-VGPMIL & $\mathbf{0.847 \pm 0.021}$ & $\mathbf{0.8216 \pm 0.0297}$ & $\mathbf{0.908 \pm 0.0174}$ \\
			&  & VGPMIL & $0.8221 \pm 0.0279$ & $0.8044 \pm 0.0328$ & $0.903 \pm 0.018$ \\
			\cline{2-6}
			& \multirow[c]{2}{*}{100} & G-VGPMIL & $\mathbf{0.8448 \pm 0.017}$ & $\mathbf{0.8207 \pm 0.0307}$ & $\mathbf{0.9113 \pm 0.0185}$ \\
			&  & VGPMIL & $0.829 \pm 0.0237$ & $0.8096 \pm 0.03$ & $0.9029 \pm 0.0181$ \\
			\cline{2-6}
			& \multirow[c]{2}{*}{200} & G-VGPMIL & $\underline{\mathbf{0.8485 \pm 0.019}}$ & $\underline{\mathbf{0.823 \pm 0.0321}}$ & $\mathbf{0.9113 \pm 0.0185}$ \\
			&  & VGPMIL & $0.8327 \pm 0.0235$ & $0.8114 \pm 0.0306$ & $0.9039 \pm 0.018$ \\
			\bottomrule
		\end{tabular}
	\end{adjustbox}
	\caption{\centering CQ500}
	\label{table:cq500}
\end{table}

\end{appendices}

\end{document}